%% file: main.tex
\documentclass[runningheads]{llncs}
\pdfoutput=1
\usepackage{graphicx}
\usepackage{comment}
\usepackage{subfigure}
\usepackage{amsmath,amssymb} 
\usepackage{color}
\usepackage{wrapfig}
\usepackage{cite}
\usepackage{listings}
\usepackage{xcolor}

\definecolor{codegreen}{rgb}{0,0.6,0}
\definecolor{codegray}{rgb}{0.5,0.5,0.5}
\definecolor{codepurple}{rgb}{0.58,0,0.82}
\definecolor{backcolour}{rgb}{0.95,0.95,0.92}

\lstdefinestyle{mystyle}{
    backgroundcolor=\color{backcolour},   
    commentstyle=\color{codegreen},
    keywordstyle=\color{magenta},
    numberstyle=\tiny\color{codegray},
    stringstyle=\color{codepurple},
    basicstyle=\ttfamily\footnotesize,
    breakatwhitespace=false,         
    breaklines=true,                 
    captionpos=b,                    
    keepspaces=true,                 
    numbers=left,                    
    numbersep=5pt,                  
    showspaces=false,                
    showstringspaces=false,
    showtabs=false,                  
    tabsize=2
}

\newcommand{\beginsupp}{%
        \setcounter{table}{0}
        \renewcommand{\thetable}{S\arabic{table}}%
        \setcounter{figure}{0}
        \renewcommand{\thefigure}{S\arabic{figure}}%
     }

\usepackage{booktabs}
\usepackage{multirow}
\usepackage{amsfonts}
\usepackage[linesnumbered,lined,vlined,ruled,commentsnumbered]{algorithm2e}
\usepackage[pagebackref=true,breaklinks=true,colorlinks,bookmarks=false]{hyperref}
\newcommand{\myparagraph}[1]{\vspace{0.1em}\noindent\textbf{#1}}
\newcommand{\mycaptionsupp}[1]{\textcolor[rgb]{0.8, 0.2, 0.06}{#1}}

\newcommand{\redtf}[1]{\textcolor[rgb]{0.8, 0.2, 0.06}{#1}}

\newcommand{\bluetf}[1]{\textcolor[rgb]{0.204,0.475,0.710}{#1}}
\newcommand{\cotronlvsapce}{\vspace{0.0cm}}
\newcommand{\cotronlvsapcea}{\vspace{0.0cm}}

\newcommand{\cotronlvsapcetablecaption}{\vspace{0.0cm}}

\begin{document}
\pagestyle{headings}
\mainmatter
\def\ECCVSubNumber{2562} 

\title{An Ensemble of Epoch-wise Empirical Bayes \\for Few-shot Learning} 

\titlerunning{An Ensemble of Epoch-wise Empirical Bayes for Few-shot Learning}
\authorrunning{Liu et al.}

\author{Yaoyao Liu$^{1}$, Bernt Schiele$^{1}$, and Qianru Sun$^{2}$\\
\tt\small{\{yaoyao.liu, schiele, qsun\}@mpi-inf.mpg.de \quad qianrusun@smu.edu.sg}
}
\institute{$^{1}$Max Planck Institute for Informatics, Saarland Informatics Campus\\
$^{2}$ School of Information Systems, Singapore Management University}

\maketitle
\begin{abstract}
\input{sections/0_abstract}    
\end{abstract}

\input{sections/1_intro.tex}
\input{sections/2_related_works.tex}
\input{sections/3_preliminary.tex}

\input{sections/4_approach.tex}
\input{sections/5_experiment.tex}

\input{sections/6_conclusion.tex}

~\\
\noindent\textbf{Acknowledgments.} This research was supported by the Singapore Ministry of Education (MOE) Academic Research Fund (AcRF) Tier 1 grant. We thank all reviewers and area chairs for their constructive suggestions.

\bibliographystyle{splncs04}
\bibliography{references}

\input{sections/7_supplementary}

\end{document}

%% file: sections/0_abstract.tex
Few-shot learning aims to train efficient predictive models with a few examples. The lack of training data leads to poor models that perform high-variance or low-confidence predictions. In this paper, we propose to meta-learn the ensemble of epoch-wise empirical Bayes models (E$^3$BM) to achieve robust predictions. ``Epoch-wise'' means that each training epoch has a Bayes model whose parameters are specifically learned and deployed. ``Empirical'' means that the hyperparameters, e.g., used for learning and ensembling the epoch-wise models, are generated by hyperprior learners conditional on task-specific data. We introduce four kinds of hyperprior learners by considering inductive \emph{vs.} transductive, and epoch-dependent \emph{vs.} epoch-independent, in the paradigm of meta-learning. We conduct extensive experiments for five-class few-shot tasks on three challenging benchmarks: \emph{mini}ImageNet, \emph{tiered}ImageNet, and FC100, and achieve top performance using the epoch-dependent transductive hyperprior learner, which captures the richest information. Our ablation study shows that both ``epoch-wise ensemble'' and ``empirical'' encourage high efficiency and robustness in the model performance\footnote{Our 
code is open-sourced at \href{https://gitlab.mpi-klsb.mpg.de/yaoyaoliu/e3bm}{https://gitlab.mpi-klsb.mpg.de/yaoyaoliu/e3bm}.}.

%% file: sections/1_intro.tex
\cotronlvsapce
\section{Introduction}
\cotronlvsapcea
\label{sec_intro}

The ability of learning new concepts from a handful of examples is well-handled by humans, while in contrast, it remains challenging for machine models whose typical training requires a significant amount of data for good performance~\cite{KrizhevskyNIPS12}.
However, in many real-world applications, we have to face the situations of lacking a significant amount of training data, as e.g., in the medical domain. It is thus desirable to improve machine learning models to handle few-shot settings where each new concept has very scarce examples~\cite{FeiFeiFP06,FinnAL17,SunCVPR2019,jung2020few}.

\input{misc/1_figure_teaser.tex}
Meta-learning methods aim to tackle the few-shot learning problem by transferring experience from similar few-shot tasks~\cite{Caruana94}. There are different meta strategies, among which the gradient descent based methods are particularly promising for today's neural networks~\cite{FinnAL17,FinnNIPS18,GrantICLR2018,FranceschiICML18,LeeICML18,ZhangNIPS2018MetaGAN,SunCVPR2019,AntoniouICLR19,zhang2019canet,hu20empirical,zhang2019pyramid,DeepEMDJ,wang2020few}. 
These methods follow a unified meta-learning procedure that contains two loops. The inner loop learns a base-learner for each individual task, and the outer loop uses the validation loss of the base-learner to optimize a meta-learner. 
In previous works~\cite{FinnAL17,FinnNIPS18,AntoniouICLR19,SunCVPR2019}, the task of the meta-learner is to initialize  the base-learner for the fast and efficient adaptation to the few training samples in the new task.

In this work, we aim to address two shortcomings of the previous works. First, 
the learning process of a base-learner for few-shot tasks is quite unstable~\cite{AntoniouICLR19}, and often results in high-variance or low-confidence predictions. 
An intuitive solution is to train an ensemble of models and use the combined prediction which should be more robust~\cite{breiman1996stacked,ozay2012new,JuBL17}.
However, it is not obvious how to obtain and combine multiple base-learners given the fact that a very limited number of training examples are available. Rather than learning multiple independent base-learners~\cite{yoon2018bayesian}, we propose a novel method of utilizing the sequence of epoch-wise base-learners (while training a single base-learner) as the ensemble.
Second, it is well-known that the values of hyperparameters, e.g., for initializing and updating models, are critical for best performance, and are particularly important for few-shot learning. In order to explore the optimal hyperparameters, we propose to employ the empirical Bayes method in the paradigm of meta-learning. In specific, we meta-learn hyperprior learners with meta-training tasks, and use them to generate task-specific hyperparameters, e.g., for updating and ensembling multiple base-learners. 
We call the resulting novel approach \textbf{E$^3$BM}, 
which learns the \textbf{E}nsemble of \textbf{E}poch-wise \textbf{E}mpirical \textbf{B}ayes \textbf{M}odels for each few-shot task.
Our ``epoch-wise models'' are \emph{different models} since each one of them is resulted from a specific training epoch and is trained with a specific set of hyperparameter values.
During test, E$^3$BM combines the ensemble of models' predictions with soft ensembling weights to produce more robust results.
In this paper, we argue that during model adaptation to the few-shot tasks, the most active adapting behaviors actually happen in the early epochs, and then converge to and even overfit to the training data in later epochs. 
Related works use the single base-learner obtained from the last epoch, so their meta-learners learn only partial adaptation experience~\cite{FinnAL17,SunCVPR2019,FinnNIPS18,hu20empirical}.
In contrast, our E$^3$BM leverages an ensemble modeling strategy that adapts base-learners at different epochs and each of them has task-specific hyperparameters for updating and ensembling. 
It thus obtains the optimized combinational adaptation experience.
Figure~\ref{teaser_figure} presents the conceptual illustration of E$^3$BM, compared to those of the classical method MAML~\cite{FinnAL17} and the state-of-the-art SIB~\cite{hu20empirical}.

\textbf{Our main contributions} are three-fold. 
(1)~A novel few-shot learning approach E$^3$BM that learns to learn and combine an ensemble of epoch-wise Bayes models for more robust few-shot learning.
(2)~Novel hyperprior learners in E$^3$BM to generate the task-specific hyperparameters for learning and combining epoch-wise Bayes models. In particular, we introduce four kinds of hyperprior learner by considering inductive~\cite{FinnAL17,SunCVPR2019} and transductive learning methods~\cite{hu20empirical}, and each with either epoch-dependent (e.g., LSTM) or epoch-independent (e.g., epoch-wise FC layer) architectures.
(3)~Extensive experiments on three challenging few-shot benchmarks, \emph{mini}ImageNet~\cite{VinyalsBLKW16}, \emph{tiered}ImageNet~\cite{RenICLR2018_semisupervised} and Fewshot-CIFAR100 (FC100)~\cite{OreshkinNIPS18}.
We plug-in our E$^3$BM to the state-of-the-art few-shot learning methods~\cite{FinnAL17,SunCVPR2019,hu20empirical} and obtain consistent performance boosts. We conduct extensive model comparison and observe that our E$^3$BM employing an epoch-dependent transductive hyperprior learner achieves the top performance on all benchmarks.

%% file: misc/1_figure_teaser.tex
\begin{figure}[t]

\includegraphics[width=\textwidth]{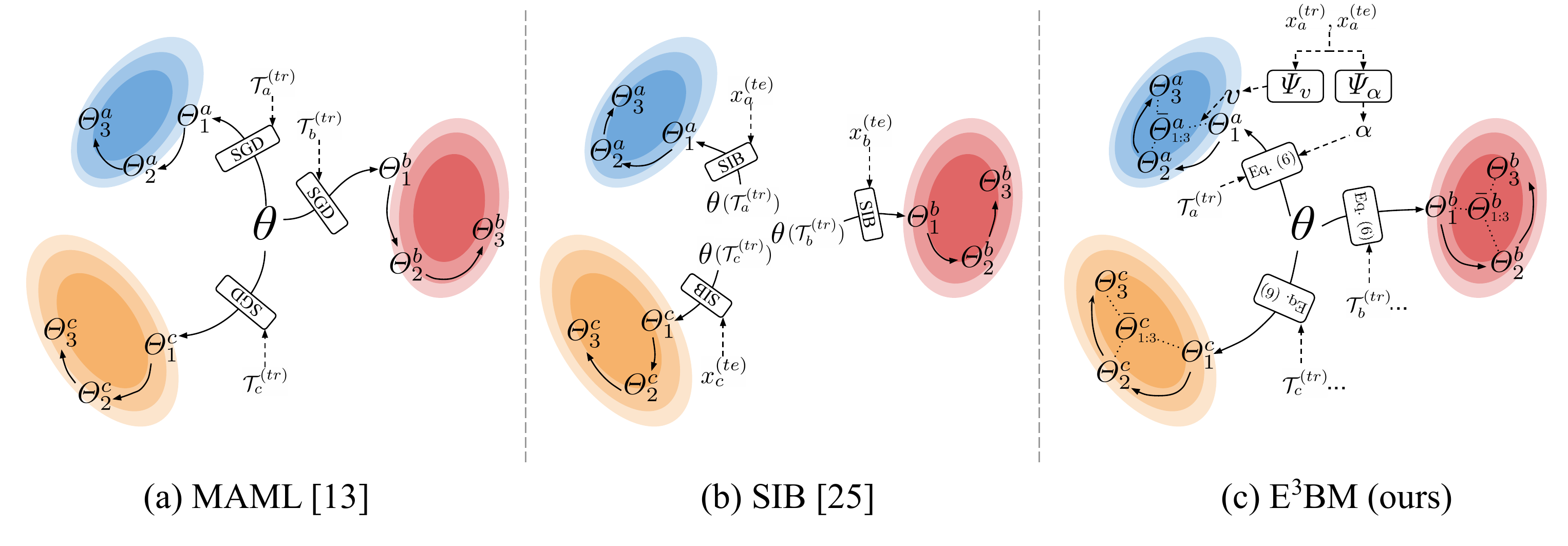}

\caption{
Conceptual illustrations of the model adaptation on the {\textcolor[rgb]{0.434, 0.656, 0.855}{\textbf{blue}}}, {\textcolor[rgb]{0.878, 0.4, 0.04}{\textbf{red}}} and {\textcolor[rgb]{0.965, 0.695, 0.42}{\textbf{yellow}}} tasks.
(a)~MAML~\cite{FinnAL17} is the classical inductive method that meta-learns a network initialization $\theta$ that is used to learn a single base-learner on each task, e.g., $\Theta^a_3$ in the blue task. (b)~SIB~\cite{hu20empirical} is a transductive method that formulates a variational posterior as a function of both labeled training data $\mathcal{T}^{(tr)}$ and unlabeled test data $x^{(te)}$. It also uses a single base-learner and optimizes the learner by running several synthetic gradient steps on $x^{(te)}$. (c) Our E$^3$BM is a generic method that learns to combine the epoch-wise base-learners (e.g., ${\Theta}_1$, ${\Theta}_2$, and ${\Theta}_3$), and to generate task-specific learning rates $\alpha$ and combination weights $v$ that encourage robust adaptation. $\bar{\Theta}_{1:3}$ denotes the ensemble result of three base-learners; $\Psi_{\alpha}$ and $\Psi_v$ denote the hyperprior learners learned to generate $\alpha$ and $v$, respectively. Note that figure (c) is based on E$^3$BM+MAML, i.e., plug-in our E$^3$BM to MAML baseline. Other plug-in versions are introduced in Sec.~\ref{subsec-plug-in}.
}
  \label{teaser_figure}
\end{figure}

%% file: sections/2_related_works.tex
\cotronlvsapce
\section{Related Works}
\cotronlvsapcea

\myparagraph{Few-shot learning \& meta-learning.}
Research literature on few-shot learning paradigms exhibits a high diversity from using data augmentation techniques~\cite{WangCVPR2018,XianCVPR2019a,ChenFZJXS19} over sharing feature representation~\cite{BartU05,WangNIPS2016} to meta-learning~\cite{Hinton1987,Thrun1998}.
In this paper, we focus on the meta-learning paradigm that leverages few-shot learning experiences from similar tasks based on the episodic formulation (see Section~\ref{sec_pre}).
Related works can be roughly divided into three categories.
(1) \emph{Metric learning methods} \cite{VinyalsBLKW16,SnellSZ17,SungCVPR2018,li2019finding,ye2018learning,LiDMMHH19,hou2019cross,dvornik2019diversity,Zhang_2020_CVPR} aim to learn a similarity space, in which the learning should be efficient for few-shot examples. The metrics include Euclidean distance~\cite{SnellSZ17}, cosine distance~\cite{VinyalsBLKW16,chen19closerfewshot}, relation module~\cite{SungCVPR2018,LiDMMHH19,hou2019cross} and graph-based similarity~\cite{SatorrasICLR2018graph,LiuICLR2019transductive}. Metric-based task-specific feature representation learning has also been presented in many related works~\cite{LiDMMHH19,ye2018learning,hou2019cross,dvornik2019diversity}.
(2) \emph{Memory network methods} \cite{MunkhdalaiICML2017,OreshkinNIPS18,MishraICLR2018} aim to learn training ``experience'' from the seen tasks and then aim to generalize to the learning of the unseen ones. A model with external memory storage is designed specifically for fast learning in a few iterations, e.g., Meta Networks~\cite{MunkhdalaiICML2017}, Neural Attentive Learner (SNAIL)~\cite{MishraICLR2018}, and Task Dependent Adaptive Metric (TADAM)~\cite{OreshkinNIPS18}.
(3) \emph{Gradient descent based methods} \cite{FinnAL17,FinnNIPS18,AntoniouICLR19,RaviICLR2017,LeeICML18,GrantICLR2018,ZhangNIPS2018MetaGAN,SunCVPR2019,lee2019meta,hu20empirical,li2019learning}
usually employ a meta-learner that learns to fast adapt an NN base-learner to a new task within a few optimization steps. 
For example, Rusu \emph{et al}.~\cite{Rusu2019} introduced a classifier generator as the meta-learner, which outputs parameters for each specific task.
Lee \emph{et al}.~\cite{lee2019meta} presented a meta-learning approach with convex base-learners for few-shot tasks. 
Finn \emph{et al}.~\cite{FinnAL17} designed a meta-learner called MAML, which learns to effectively initialize the parameters of an NN base-learner for a new task. 
Sun \emph{et al}.~\cite{SunCVPR2019,SUNCVPR2019journal} introduced an efficient knowledge transfer operator on deeper neural networks and achieved a significant improvement for few-shot learning models.
Hu \emph{et al}.~\cite{hu20empirical} proposed to update base-learner with synthetic gradients generated by a variational posterior conditional on unlabeled data.
Our approach is closely related to gradient descent based methods~\cite{FinnAL17,AntoniouICLR19,SunCVPR2019,SunCVPR2019,SUNCVPR2019journal,hu20empirical}.
An important difference is that we learn how to combine an ensemble of epoch-wise base-learners and how to generate efficient hyperparameters for base-learners, while other methods such as MAML~\cite{FinnAL17}, MAML++~\cite{AntoniouICLR19}, LEO~\cite{Rusu2019}, MTL~\cite{SunCVPR2019,SUNCVPR2019journal}, and SIB~\cite{hu20empirical} use a single base-learner.

\myparagraph{Hyperparameter optimization.}
Building a model for a new task is a process of exploration-exploitation.
Exploring suitable architectures and hyperparameters are important before training.
Traditional methods are model-free, e.g., based on grid search~\cite{BergstraIMLR12,LiICML17,Jaderberg17}. 
They require multiple full training trials and are thus costly. 
Model-based hyperparameter optimization methods are adaptive but sophisticated, e.g., using random forests~\cite{HutterLION11}, Gaussian processes~\cite{SnoekNIPS12} and input warped Gaussian processes~\cite{SnoekICML14} or scalable Bayesian optimization~\cite{SnoekICML15}. 
In our approach, we meta-learn a hyperprior learner to output optimal hyperparameters by gradient descent, without additional manual labor.
Related methods using gradient descent mostly work for single model learning in an inductive way~\cite{Bengio2000,Domke12,Maclaurin2015,Luketina2016,FranceschiICML18,MetzICLR19,LiICML2018,Liu_2020_CVPR}.
While, our hyperprior learner generates a sequence of hyperparameters for multiple models, in either the inductive or the transductive learning manner.

\myparagraph{Ensemble modeling.}
It is a strategy~\cite{huang2017snapshot,zhang2019nonlinear} to use multiple algorithms to improve machine learning performance, and which is proved to be effective to reduce the problems related to overfitting~\cite{KunchevaW03,SollichK95}.
Mitchell et al.~\cite{mitchell1997machine} provided a theoretical explanation for it. 
Boosting is one classical way to build an ensemble, e.g., AdaBoost~\cite{freund1997decision} and Gradient Tree Boosting~\cite{friedman2002stochastic}.
Stacking combines multiple models by learning a combiner and it applies to both tasks in supervised learning~\cite{breiman1996stacked,ozay2012new,JuBL17} and unsupervised learning~\cite{smyth1999linearly}.
Bootstrap aggregating (i.e., Bagging) builds an ensemble of models through parallel training~\cite{breiman1996stacked}, 
e.g., random forests~\cite{ho1995random}. 
The ensemble can also be built on a temporal sequence of models~\cite{LaineICLR17}. 
Some recent works have applied ensemble modeling to few-shot learning.
Yoon et al. proposed Bayesian MAML (BMAML) that trains multiple instances of base-model to reduce mete-level overfitting~\cite{YoonKDKBA18}. 
The most recent work~\cite{DiversityICCV2019} encourages multiple networks to cooperate while keeping predictive diversity. Its networks are trained with carefully-designed penalty functions, different from our automated method using empirical Bayes. Besides, its method needs to train much more network parameters than ours. Detailed comparisons are given in the experiment section.

%% file: sections/3_preliminary.tex
\section{Preliminary}
\label{sec_pre}
In this section, we introduce the unified episodic formulation of few-shot learning, following~\cite{VinyalsBLKW16,RaviICLR2017,FinnAL17}. 
This formulation was proposed for few-shot classification first in~\cite{VinyalsBLKW16}.
Its problem definition is different from traditional classification in three aspects: (1)~the main phases are not training and test but meta-training and meta-test, each of which includes training and test;
(2)~the samples in meta-training and meta-testing are not datapoints but episodes, i.e. few-shot classification tasks; and (3)~the objective is not classifying unseen datapoints but to fast adapt the meta-learned knowledge to the learning of new tasks.

Given a dataset $\mathcal{D}$ for meta-training, we first sample few-shot episodes (tasks) $\{\mathcal{T}\}$ from a task distribution $p(\mathcal{T})$ such that each episode $\mathcal{T}$ contains a few samples of a few classes, e.g., $5$ classes and $1$ shot per class. 
Each episode $\mathcal{T}$ includes a training split $\mathcal{T}^{(tr)}$ to optimize a specific base-learner, and a test split $\mathcal{T}^{(te)}$ to compute a generalization loss to optimize a global meta-learner.
For meta-test, given an unseen dataset $\mathcal{D}_{un}$ (i.e., samples are from unseen classes), we sample a test task $\mathcal{T}_{un}$ to have the same-size training/test splits. 
We first initiate a new model with meta-learned network parameters (output from our hyperprior learner), then train this model on the training split $\mathcal{T}^{(tr)}_{un}$.
We finally evaluate the performance on the test split $\mathcal{T}^{(te)}_{un}$. 
If we have multiple tasks, we report average accuracy as the final result.

%% file: sections/4_approach.tex
\cotronlvsapce
\section{An Ensemble of Epoch-wise Empirical Bayes Models 
}
\cotronlvsapcea
\label{sec_method}
\input{misc/4_sub_figure_framework.tex}

As shown in Fig.\thinspace\ref{figure_netarch}, E$^3$BM trains a sequence of epoch-wise base-learners $\{\Theta_m\}$ with training data $\mathcal{T}^{(tr)}$ and learns to combine their predictions $\{z^{(te)}_m\}$ on test data $x^{(te)}$ for the best performance. This ensembling strategy achieves more robustness during prediction. The hyperparameters of each base-learner, i.e.,
learning rates $\alpha$ and combination weights $v$, are generated by the hyperprior learners conditional on task-specific data, e.g., $x^{(tr)}$ and $x^{(te)}$. This approach encourages the high diversity and informativeness of the ensembling models.

\subsection{Empirical Bayes method}
\label{sec-empirical-bayes}

Our approach can be formulated as an empirical Bayes method that learns two levels 
of models for a few-shot task. The first level has hyperprior learners that generate hyperparameters for updating and combining the second-level models. More specifically, these second-level models are trained with the loss derived from the combination of their predictions on training data. After that, their loss of test data are used to optimize the hyperprior learners. This process is also called meta update, see the dashed arrows in Fig.~\ref{figure_netarch}.

In specific, we sample $K$ episodes $\{\mathcal{T}_k\}_{k=1}^K$ from the meta-training data $\mathcal{D}$. Let $\Theta$ denote base-learner and $\psi$ represent its hyperparameters. 
An episode $\mathcal{T}_k$ aims to train $\Theta$ to recognize different concepts, so we consider to use concepts related (task specific) data for customizing the $\Theta$
through a hyperprior $p(\psi_k)$.
To achieve this, we first formulate the empirical Bayes method with marginal likelihood according to hierarchical structure among data as follows,
\begin{equation} 
    \label{eq_hierarchical_bayes1}
    p(\mathcal{T}) = \prod_{k=1}^K p(\mathcal{T}_k) 
    = \prod_{k=1}^K \int_{\psi_k} p(\mathcal{T}_k|\psi_k)p(\psi_k)d{\psi_k}.
\end{equation}

Then, we use variational inference~\cite{hoffman2013stochastic} to estimate $\{p(\psi_k)\}_{k=1}^K$. 
We parametrize distribution $q_{\varphi_k}(\psi_k)$ with $\varphi_k$ for each $p(\psi_k)$, and update $\varphi_k$ to increase the similarity betweeen $q_{\varphi_k}(\psi_k)$ and $p(\psi_k)$. 
As in standard probabilistic modeling, we derive an evidence lower bound on the log version of Eq.~\eqref{eq_hierarchical_bayes1} to update $\varphi_k$,
\begin{equation} 
    \label{eq_hierarchical_bayes_log_elbo}
    \log p(\mathcal{T}) \geqslant \sum_{k=1}^K\Big[ \mathbb{E}_{\psi_k\sim q_{\varphi_k}} \big[ \log p(\mathcal{T}_k|\psi_k) \big] - D_{\mathrm{KL}}(q_{\varphi_k}(\psi_k)||p(\psi_k))\Big].
\end{equation}

Therefore, the problem of using $q_{\varphi_k}(\psi_k)$ to approach to the best estimation of $p(\psi_k)$ 
becomes equivalent to the objective of
maximizing the evidence lower bound~\cite{blei2017variational,hoffman2013stochastic,hu20empirical} in Eq.~\eqref{eq_hierarchical_bayes_log_elbo}, with respect to $\{\varphi_k\}_{k=1}^K$, as follows,
\begin{equation} 
    \label{eq_hierarchical_bayes_optimization_problem}
    \min_{\{\varphi_k\}_{k=1}^K} \frac{1}{K} \sum_{k=1}^K\Big[ \mathbb{E}_{\psi_k\sim q_{\varphi_k}} \big[ -\log p(\mathcal{T}_k|\psi_k) \big] + D_{\mathrm{KL}}(q_{\varphi_k}(\psi_k)||p(\psi_k))\Big].
\end{equation}

\input{misc/4_sub_figure_framework_hyperprior_learner}
To improve the robustness of few-shot models, existing methods sample a significant amount number of episodes during meta-training \cite{FinnAL17,SunCVPR2019}. 
Each episode employing
its own hyperprior $p(\psi_k)$ causes a huge computation burden, making it difficult to solve the aforementioned optimization problem.
To tackle this, we leverage a technique called ``amortized variational inference'' \cite{kingma2013auto,RezendeMW14,hu20empirical}.
We parameterize the KL term in $\{\varphi_k\}_{k=1}^K$ (see Eq.~\eqref{eq_hierarchical_bayes_optimization_problem}) with a unified deep neural network $\Psi(\cdot)$ taking $x^{(tr)}_k$ (inductive learning) or $\{x^{(tr)}_k, x^{(te)}_k\}$ (transductive learning) as inputs, where $x^{(tr)}_k$ and $x^{(te)}_k$ respectively denote the training and test samples in the $k$-th episode. In this paper, we call $\Psi(\cdot)$ hyperprior learner.
As shown in Fig.\thinspace\ref{figure_netarch_hyper}, we additionally feed the hyperprior learner with the training gradients $\nabla\mathcal{L}_{\Theta}(\mathcal{T}^{(tr)}_k)$ to $\Psi(\cdot)$ to encourage it to ``consider'' the current state of the training epoch. 
We mentioned in Sec.~\ref{sec_intro} that base-learners at different epochs are adapted differently, so we expect the corresponding hyperprior learner to ``observe'' and ``utilize'' this information to produce effective hyperparameters.
By replacing $q_{\varphi_k}$ with $q_{\Psi(\cdot)}$, Problem~\eqref{eq_hierarchical_bayes_optimization_problem} can be rewritten as:
\begin{equation} 
    \label{eq_hierarchical_bayes_optimization_problem_new}
    \min_{\Psi} \frac{1}{K} \sum_{k=1}^K\Big[ \mathbb{E}_{\psi_k\sim q_{\Psi(\cdot)}} \big[ -\log p(\mathcal{T}_k|\psi_k) \big] + D_{\mathrm{KL}}(q_{\Psi(\cdot)}(\psi_k)||p(\psi_k))\Big].
\end{equation}
Then, we solve Problem~\eqref{eq_hierarchical_bayes_optimization_problem_new} by optimizing $\Psi(\cdot)$ with the meta gradient descent method used in classical meta-learning paradigms~\cite{FinnAL17,SunCVPR2019,hu20empirical}. We elaborate the details of learning $\{\Theta_m\}$ and meta-learning $\Psi(\cdot)$ in the following sections.

\subsection{Learning the ensemble of base-learners}
\label{sec-ensemble}

Previous works have shown that training multiple instances of the base-learner is helpful to achieve robust few-shot learning~\cite{yoon2018bayesian,dvornik2019diversity}. However, they suffer from the computational burden of optimizing multiple copies of neural networks in parallel, and are not easy to generalize to deeper neural architectures. If include the computation of second-order derivatives in meta gradient descent~\cite{FinnAL17}, this burden becomes more unaffordable.
In contrast, our approach is free from this problem, because it is built on top of optimization-based meta-learning models, e.g., MAML~\cite{FinnAL17}, MTL~\cite{SunCVPR2019}, and SIB~\cite{hu20empirical}, which naturally produce a sequence of models along the training epochs in each episode. 

Given an episode $\mathcal{T}=\{\mathcal{T}^{(tr)}, \mathcal{T}^{(te)}\}=\{\{x^{(tr)}, y^{(tr)}\},\{x^{(te)}, y^{(te)}\} \}$,
let $\Theta_{m}$ denote the parameters of the base-learner working at epoch $m$ (w.r.t. $m$-th base-learner or BL-$m$), with $m \in \{1, ..., M\}$.
Basically, we initiate BL-$1$ with parameters $\theta$ (network weights and bias) and hyperparameters (e.g., learning rate $\alpha$), where $\theta$ is meta-optimized as in MAML~\cite{FinnAL17}, and $\alpha$ is generated by the proposed hyperprior learner $\Psi_{\alpha}$.
We then adapt BL-$1$ with normal gradient descent on the training set $\mathcal{T}^{(tr)}$, and use the adapted weights and bias to initialize BL-$2$. 
The general process is thus as follows,
\begin{equation} 
\label{eq_theta_ini}
\Theta_{0} \gets \theta,
\end{equation}
\begin{equation} 
    \label{eq_theta_normal}
    \Theta_{m} \gets \Theta_{m-1} - \alpha_{m}\nabla_{\Theta}\mathcal{L}^{(tr)}_{m} = \Theta_{m-1} - \Psi_{\alpha}(\tau, \nabla_{\Theta}\mathcal{L}^{(tr)}_{m})\nabla_{\Theta}\mathcal{L}^{(tr)}_{m},
\end{equation}
where $\alpha_{m}$ is the learning rate outputted from $\Psi_{\alpha}$, and $\nabla_{\Theta}\mathcal{L}^{(tr)}_{m}$ are the derivatives of the training loss, i.e, gradients. $\tau$ represents either $x^{(tr)}$ in the inductive setting, or $\{x^{(tr)}, x^{(te)}\}$ in the transductive setting.
Note that $\Theta_0$ is introduced to make the notation consistent, and a subscript $m$ is omitted from $\Psi_{\alpha}$ for conciseness. 
Let $F(x; \Theta_m)$ denote the prediction scores of input $x$,
so the base-training loss $\mathcal{T}^{(tr)}=\big\{x^{(tr)}, y^{(tr)}\big\}$ can be unfolded as,
\begin{equation} 
    \label{eq_base_loss_normal}
    \mathcal{L}^{(tr)}_{m}= L_{ce}\big(F(x^{(tr)}; \Theta_{m-1}), y^{(tr)}\big),
\end{equation}
where $L_{ce}$ is the softmax cross entropy loss. During episode test,
each base-learner BL-$m$ infers the prediction scores $z_m$ for test samples $x^{(te)}$,
\begin{equation}
    \label{MAML_pred}
    z_m = F(x^{(te)}; \Theta_m).
\end{equation}

Assume the hyperprior learner $\Psi_{v}$ generates the combination weight $v_m$ for BL-$m$.
The final prediction score is initialized as $\hat{y}^{(te)}_1=v_1 z_1$ .
For the $m$-th base epoch, the prediction $z_m$ will be calculated and added to $\hat{y}^{(te)}$ as follows, 
\begin{equation}
    \label{yhat_update}
    \hat{y}^{(te)}_m \gets v_m z_m + \hat{y}^{(te)}_{m-1} = \Psi_{v}(\tau, \nabla_{\Theta}\mathcal{L}^{(tr)}_{m}) F(x^{(te)}; \Theta_{m}) + \hat{y}^{(te)}_{m-1}.
\end{equation}
In this way, we can update prediction scores without storing base-learners or feature maps in the memory.

\subsection{Meta-learning the hyperprior learners}
\label{subsec_hyperprior}

As presented in Fig.\thinspace\ref{figure_netarch_hyper}, we introduce two architectures, i.e., LSTM or individual FC layers, for the hyperprior learner. 
FC layers at different epochs are independent.
Using LSTM to ``connect'' all epochs is expected to ``grasp'' more task-specific information from the overall training states of the task. In the following, we elaborate the meta-learning details for both designs.

Assume before the $k$-th episode, we have meta-learned the base learning rates $\{\alpha_m'\}_{m=1}^M$ and combination weights $\{v_m'\}_{m=1}^M$.
Next in the $k$-th episode, specifically at the $m$-th epoch as shown in Fig.\thinspace\ref{figure_netarch_hyper},
we compute the mean values of $\tau$ and
$\nabla_{\Theta_m}\mathcal{L}^{(tr)}_m$, respectively, over all samples\footnote{In the inductive setting, training images are used to compute $\bar{\tau}$; while in the transductive setting, test images are additionally used.}.
We then input the concatenated value to FC or LSTM mapping function as follows,
\begin{equation}
    \label{eq_psi_flow_fc}
    \Delta\alpha_m, \Delta v_m = \textrm{FC}_m(\textrm{concat}[\bar{\tau}; \overline{\nabla_{\Theta_m}\mathcal{L}^{(tr)}_m}]), \ \textbf{\textrm{or}}
\end{equation}
\begin{equation}
    \label{eq_psi_flow_L}
    [\Delta\alpha_m, \Delta v_m], h_{m} = \textrm{LSTM}(\textrm{concat}[\bar{\tau}; \overline{\nabla_{\Theta_m}\mathcal{L}^{(tr)}_m}], h_{m-1}),
\end{equation}
where $h_{m}$ and $h_{m-1}$ are the hidden states at epoch $m$ and epoch $m-1$, respectively. 
We then use the output values to update hyperparameters as,
\begin{equation}
    \label{eq_psi_final}
    \alpha_m = \lambda_1\alpha_m' + (1-\lambda_1)\Delta\alpha, \ v_m = \lambda_2 v_m' + (1-\lambda_2)\Delta v,
\end{equation}
where $\lambda_1$ and $\lambda_2$ are fixed fractions in $(0, 1)$.
Using learning rate $\alpha_m$, we update BL-$(m-1)$ to be BL-$m$ with Eq.~\eqref{eq_theta_normal}. 
After $M$ epochs, we obtain the combination of predictions $\hat{y}^{(te)}_M$ (see Eq.~\eqref{yhat_update}) on test samples. In training tasks, we compute the test loss as,
\begin{equation} 
    \label{eq_epite_loss_weighted}
    \mathcal{L}^{(te)}={L}_{ce}(\hat{y}^{(te)}_M ,y^{(te)}).
\end{equation}
We use this loss to calculate meta gradients to update $\Psi$ as follows,
\begin{equation} 
    \label{eq_update_alpha}
      \Psi_{\alpha} \gets \Psi_{\alpha} - \beta_1\nabla_{\Psi_{\alpha}}\mathcal{L}^{(te)}, \ \  \Psi_{v} \gets \Psi_{v} - \beta_2\nabla_{\Psi_{v}}\mathcal{L}^{(te)},
\end{equation}
where $\beta_1$ and $\beta_2$ are meta-learning rates that determine the respective stepsizes for updating $\Psi_{\alpha}$ and $\Psi_{v}$.
These updates are to back-propagate the test gradients till the input layer, through unrolling all base training gradients of $\Theta_1\sim \Theta_M$. The process thus involves a gradient through a gradient~\cite{FinnAL17,FinnNIPS18,SunCVPR2019}. Computationally, it requires an additional backward pass through $\mathcal{L}^{(tr)}$ to compute Hessian-vector products, which is supported by standard numerical computation libraries such as TensorFlow~\cite{girija2016tensorflow} and PyTorch~\cite{paszke2019pytorch}.


\subsection{Plugging-in E$^3$BM to baseline methods}
\label{subsec-plug-in}
The optimization of $\Psi$ relies on meta gradient descent method which was first applied to few-shot learning in MAML~\cite{FinnAL17}.
Recently, MTL~\cite{SunCVPR2019} showed more efficiency by implementing that method on deeper pre-trained CNNs (e.g., ResNet-12~\cite{SunCVPR2019}, and ResNet-25~\cite{SUNCVPR2019journal}).
SIB~\cite{hu20empirical} was built on even deeper and wider networks (WRN-28-10), and it achieved top performance by synthesizing gradients in transductive learning.
These three methods are all optimization-based, and use the single base-learner of the last base-training epoch. 
In the following, we describe how to learn and combine multiple base-learners in MTL, SIB and MAML, respectively, using our E$^3$BM approach.

According to~\cite{SunCVPR2019,hu20empirical}, we pre-train the feature extractor $f$ on a many-shot classification task using the whole set of $\mathcal{D}$. The meta-learner in MTL is called scaling and shifting weights $\Phi_{SS}$, and in SIB is called synthetic information bottleneck network $\phi(\lambda, \xi)$. 
Besides, there is a common meta-learner called base-learner initializer $\theta$, i.e., the same $\theta$ in Fig.\thinspace\ref{figure_netarch}, in both methods. In MAML, the only base-learner is $\theta$ and there is no pre-training for its feature extractor $f$.

Given an episode $\mathcal{T}$, we feed training images $x^{(tr)}$ and test images $x^{(te)}$ to the feature extractor $f\odot\Phi_{SS}$ in MTL ($f$ in SIB and MAML), and obtain the embedding $e^{(tr)}$ and $e^{(te)}$, respectively.
Then in MTL, we use $e^{(tr)}$ with labels to train base-learner $\Theta$ for $M$ times to get $\{\Theta_m\}_{m=1}^M$ with Eq.~\eqref{eq_theta_normal}. In SIB, we use its multilayer perceptron (MLP) net to synthesize gradients conditional on $e^{(te)}$ to indirectly update $\{\Theta_m\}_{m=1}^M$. During these updates, our hyperprior learner $\Psi_{\alpha}$ derives the learning rates for all epochs.
In episode test, we feed $e^{(te)}$ to $\{\Theta_m\}_{m=1}^M$ and get the combined prediction $\{z_m\}_{m=1}^M$ with Eq.~\eqref{yhat_update}. 
Finally, we compute the test loss to meta-update $[\Psi_{\alpha}; \Psi_{v}; \Phi_{SS}; \theta]$ in MTL, $[\Psi_{\alpha}; \Psi_{v}; \phi(\lambda, \xi); \theta]$ in SIB, and $[f; \theta]$ in MAML. We call the resulting methods MTL+E$^3$BM, SIB+E$^3$BM, and MAML+E$^3$BM, respectively, and demonstrate their improved efficiency over baseline models~\cite{SunCVPR2019,hu20empirical,FinnAL17} in experiments.

%% file: misc/4_sub_figure_framework.tex
\begin{figure*}[t]
\centering
\includegraphics[width=\textwidth]{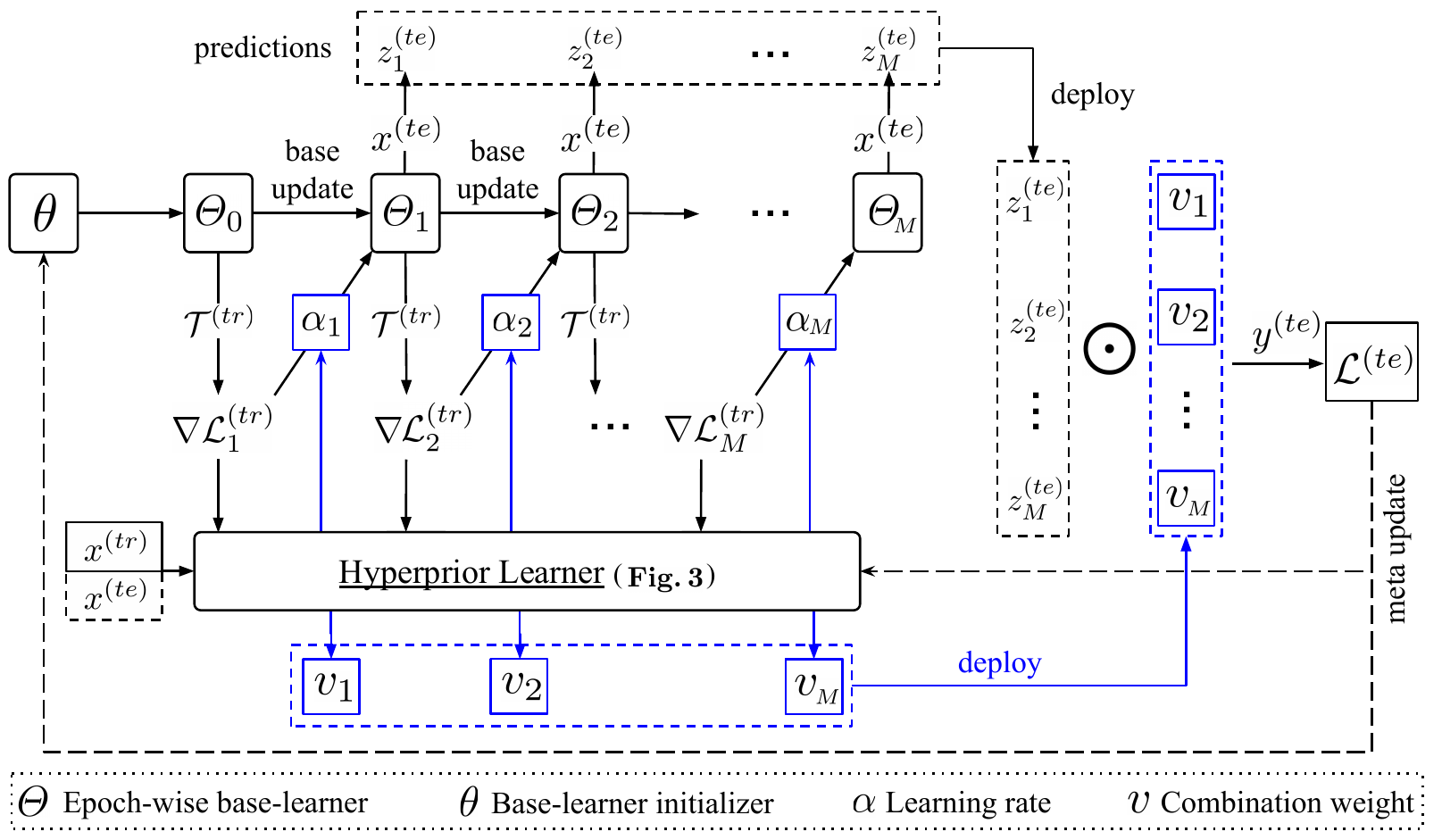}
\caption{The computing flow of the proposed E$^3$BM approach in one meta-training episode. For the meta-test task, the computation will be ended with predictions. Hyper-learner predicts task-specific hyperparameters, i.e., learning rates and multi-model combination weights. When its input contains $x^{(te)}$, it is transductive, otherwise inductive. Its detailed architecture is given in Fig.~\ref{figure_netarch_hyper}.}
\label{figure_netarch}

\end{figure*}

%% file: misc/4_sub_figure_framework_hyperprior_learner.tex
\begin{wrapfigure}{r}{0.57\textwidth}
\centering

\includegraphics[width=0.55\textwidth]{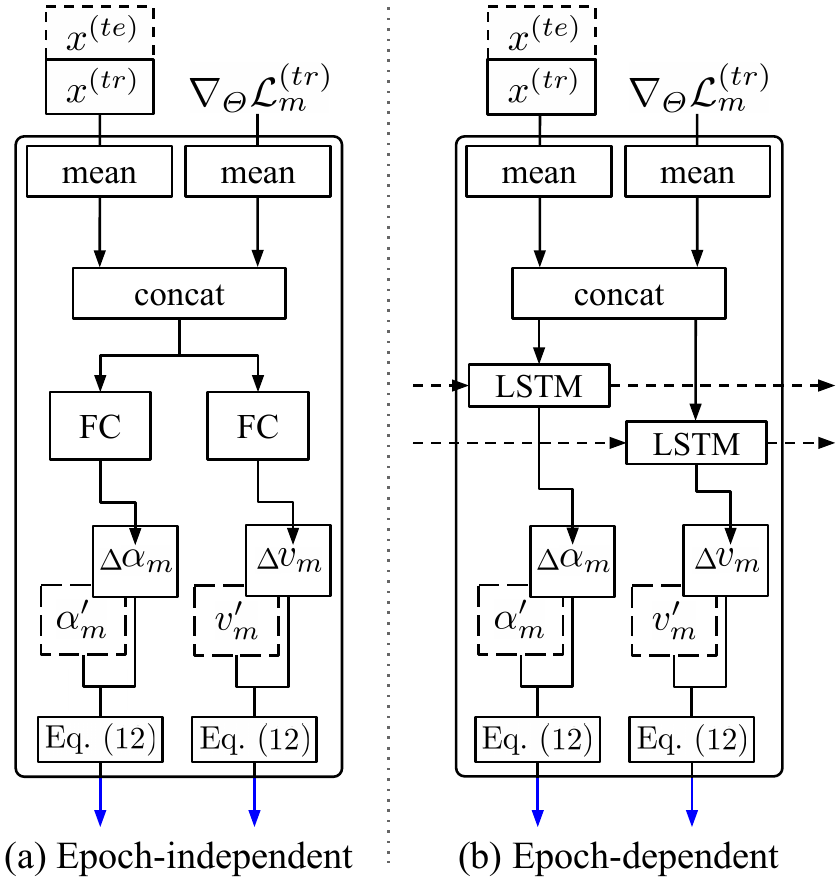}
\caption{Two options of hyperprior learner at the $m$-th base update epoch. In terms of the mapping function, we deploy either FC layers to build epoch-independent hyperprior learners, or LSTM to build an epoch-dependent learner. Values in dashed box were learned from previous tasks.}
\label{figure_netarch_hyper}

\end{wrapfigure}

%% file: sections/5_experiment.tex
\cotronlvsapce
\section{Experiments}
\cotronlvsapcea
\label{sec_exp}

We evaluate our approach in terms of its overall performance and the effects of its two components, i.e. ensembling epoch-wise models and meta-learning hyperprior learners.
In the following sections, we introduce the datasets and implementation details, compare our best results to the state-of-the-art, and conduct an ablation study.

\subsection{Datasets and implementation details}
\label{sec_dataset}

\myparagraph{Datasets.} We conduct few-shot image classification experiments on three benchmarks: \emph{mini}ImageNet~\cite{VinyalsBLKW16}, \emph{tiered}ImageNet~\cite{RenICLR2018_semisupervised} and FC100~\cite{OreshkinNIPS18}.
\emph{mini}ImageNet is the most widely used in related works~\cite{FinnAL17,hu20empirical,hou2019cross,SunCVPR2019,SungCVPR2018,hu20empirical}. \emph{tiered}ImageNet and FC100 are either with a larger scale or a more challenging setting with lower image resolution, and have stricter training-test splits.

\myparagraph{\emph{mini}ImageNet} was proposed in~\cite{VinyalsBLKW16} based on ImageNet~\cite{Russakovsky2015}.
There are $100$ classes with $600$ samples 
per class.
Classes are divided into $64$, $16$, and $20$ classes respectively for sampling tasks for meta-training, meta-validation and meta-test.
\myparagraph{\emph{tiered}ImageNet} was proposed in~\cite{RenICLR2018_semisupervised}. It contains a larger subset of ImageNet~\cite{Russakovsky2015} with $608$ classes ($779,165$ images) grouped into $34$ super-class nodes. These nodes are partitioned into $20$, $6$, and $8$ disjoint sets respectively for meta-training, meta-validation and meta-test. 
Its super-class based training-test split results in a more challenging and realistic regime with test tasks that are less similar to training tasks.
\myparagraph{FC100}
is based on the CIFAR100~\cite{CIFAR100}. The few-shot task splits were proposed in~\cite{OreshkinNIPS18}.
It contains $100$ object classes and each class has $600$ samples of $32 \times 32$ color images per class.
On these datasets, we consider the ($5$-class, $1$-way) and ($5$-class, $5$-way) classification tasks. We use the same task sampling strategy as in related works~\cite{FinnAL17,AntoniouICLR19,hu20empirical}.

\myparagraph{Backbone architectures.} 
In MAML+E$^3$BM, we use a $4$-layer convolution network (4CONV)~\cite{FinnAL17,AntoniouICLR19}.
In MTL+E$^3$BM, we use a $25$-layer residual network (ResNet-25)~\cite{QiaoCVPR2018,ye2018learning,SUNCVPR2019journal}. Followed by convolution layers, we apply an average pooling layer and a fully-connected layer.
In SIB+E$^3$BM, we use a $28$-layer wide residual network (WRN-28-10) as SIB~\cite{hu20empirical}.

\myparagraph{The configuration of base-learners.} In MTL~\cite{SunCVPR2019} and SIB~\cite{hu20empirical}, the base-learner is a single fully-connected layer. In MAML~\cite{FinnAL17}, the base-learner is the $4$-layer convolution network. In MTL and MAML, the base-learner is randomly initialized and updated during meta-learning. In SIB, the base-learner is initialized with the averaged image features of each class. The number of base-learners $M$ in MTL+E$^3$BM and SIB+E$^3$BM are respectively $100$ and $3$, i.e., the original numbers of training epochs in~\cite{SunCVPR2019,hu20empirical}.

\myparagraph{The configuration of hyperprior learners.} 
In Fig.\thinspace\ref{figure_netarch_hyper}, we show two options for hyperprior learners (i.e., $\Psi_{\alpha}$ and $\Psi_v$). Fig.\thinspace\ref{figure_netarch_hyper}(a) is the epoch-independent option, where each epoch has two FC layers to produce $\alpha$ and $v$ respectively. Fig.\thinspace\ref{figure_netarch_hyper}(b) is the epoch-dependent option which uses an LSTM to generate $\alpha$ and $v$ at all epochs.
In terms of the learning hyperprior learners, we have two settings: inductive learning denoted as ``Ind.'', and transductive learning as ``Tra.''. 
``Ind.'' is the supervised learning in classical few-shot learning methods~\cite{FinnAL17,SunCVPR2019,lee2019meta,VinyalsBLKW16,SnellSZ17}. ``Tra.'' is semi-supervised learning, based on the assumption that all test images of the episode are available. It has been applied to many recent works~\cite{LiuICLR2019transductive,hou2019cross,hu20empirical}.

\myparagraph{Ablation settings.}
We conduct a careful ablative study for two components, i.e., ``ensembling multiple base-learners'' and ``meta-learning hyperprior learners''. We show their effects indirectly by comparing our results to those of using arbitrary constant or learned values of $v$ and $\alpha$. 
\myparagraph{In terms of $v$}, we have $5$ ablation options: (v1) ``E$^3$BM'' is our method generating $v$ from $\Psi_{v}$; (v2) ``learnable'' is to set $v$ to be update by meta gradient descent same as $\theta$ in~\cite{FinnAL17}; (v3) ``optimal'' means using the values learned by option (a2) and freezing them during the actual learning; (v4) ``equal'' is an simple baseline using equal weights; (v5) ``last-epoch'' uses only the last-epoch base-learner, i.e., $v$ is set to $[0,0,...,1]$. In the experiments of (v1)-(v5), we simply set  $\alpha$ as in the following (a4)~\cite{FinnAL17,SunCVPR2019,hu20empirical}. 
\myparagraph{In terms of $\alpha$}, we have $4$ ablation options: (a1) ``E$^3$BM'' is our method generating $\alpha$ from $\Psi_{\alpha}$; (a2) ``learnable'' is to set $\alpha$ to be update by meta gradient descent same as $\theta$ in~\cite{FinnAL17}; (a3) ``optimal'' means using the values learned by option (a2) and freezing them during the actual learning; (a4) ``fixed'' is a simple baseline that uses manually chosen $\alpha$ following~\cite{FinnAL17,SunCVPR2019,hu20empirical}.
In the experiments of (a1)-(a4), we simply set $v$ as in (v5), same with the baseline method~\cite{SunCVPR2019}. 

\subsection{Results and analyses}
\label{subsec_results_and_analyses}

\input{tables/5_table_sota_new.tex}
\input{tables/5_table_e3bm_ablation_new}

In Table~\ref{tab_e3bm_sota}, we compare our best results to the state-of-the-arts. 
In Table~\ref{tab_e3bm_ablation}, we present the results of using different kinds of hyperprior learner, i.e., regarding two architectures (FC and LSTM) and two learning strategies (inductive and transductive).
In Fig.\thinspace\ref{plot_mini_val_acc}(a)(b), we show the validation results of our ablative methods, and demonstrate the change during meta-training iterations.
In Fig.\thinspace\ref{plot_mini_val_acc}(c)(d), we plot the generated values of $v$ and $\alpha$ during meta-training.

\myparagraph{Comparing to the state-of-the-arts.} Table~\ref{tab_e3bm_sota} shows that the proposed E$^3$BM achieves the best few-shot classification performance in both $1$-shot and $5$-shot settings, on three benchmarks. Please note that \cite{dvornik2019diversity} reports the results of using different backbones and input image sizes. We choose its results under the same setting of ours, i.e., using WRN-28-10 networks and $80\times80\times3$ images, for fair comparison. 
In our approach, plugging-in E$^3$BM to the state-of-the-art model SIB achieves $1.6\%$ of improvement on average, based on the identical network architecture. This improvement is significantly larger as $2.9\%$ when taking MAML as baseline.
All these show to be more impressive if considering the tiny overheads from pluging-in. For example, using E$^3$BM adds only $0.04\%$ learning parameters to the original SIB model, and it gains only $5.2\%$ average overhead regarding the computational time.
It is worth mentioning that the amount of learnable parameters in SIB+E$^3$BM is around $80\%$ less than that of model in \cite{dvornik2019diversity} which ensembles $5$ deep networks in parallel (and later learns a distillation network).

\myparagraph{Hyperprior learners.} 
In Table~\ref{tab_e3bm_ablation}, we can see that using transductive learning clearly outperforms inductive learning, e.g., No. 5 \emph{vs.} No. 4.
This is because the ``transduction'' leverages additional data, i.e., the episode-test images (no labels), during the base-training. %
In terms of the network architecture, we observe that LSTM-based learners are slightly better than FC-based (e.g., No. 3 \emph{vs.} No. 2). LSTM is a
sequential model and is indeed able to ``observe'' more patterns from the adaptation behaviors of models at adjacent epochs.

\myparagraph{Ablation study.} 
Fig.\thinspace\ref{plot_mini_val_acc}(a) shows the comparisons among $\alpha$ related ablation models. 
Our E$^3$BM ({\textcolor[rgb]{0.933,0.467,0.318}{\textbf{orange}}}) again performs the best, over the models of using any arbitrary $\alpha$ ({\textcolor[rgb]{0.8, 0.2, 0.062}{\textbf{red}}} or {\textcolor[rgb]{0.2, 0.73, 0.93}{\textbf{light blue}}}),
as well as over the model
with $\alpha$ optimized by the meta gradient descent ({\textcolor[rgb]{0.204, 0.475, 0.710}{\textbf{blue}}})~\cite{FinnAL17}.
Fig.\thinspace\ref{plot_mini_val_acc}(b)
shows that our approach E$^3$BM
works 
consistently better than the ablation models related to $v$.
We should emphasize that E$^3$BM is clearly more efficient than 
the model trained with meta-learned $v$ ({\textcolor[rgb]{0.204, 0.475, 0.710}{\textbf{blue}}}) through meta gradient descent~\cite{FinnAL17}. 
This is because E$^3$BM hyperprior learners generate empirical weights conditional on task-specific data. 
The LSTM-based learners can leverage even more task-specific information, i.e., the hidden states from previous epochs, to improve the efficiency.

\input{misc/5_mini_1shot_valacc}

\myparagraph{The values of $\alpha$ and $v$ learned by E$^3$BM.} 
Fig.\thinspace\ref{plot_mini_val_acc}(c)(d) shows the values of $\alpha$ and $v$ during the meta-training iterations in our approach.
Fig.\thinspace\ref{plot_mini_val_acc}(c) show the base-learners working at later training epochs (e.g., BL-100) tend to get smaller values of $\alpha$. This is actually similar to the common manual schedule, i.e. monotonically decreasing learning rates, of conventional large-scale network training~\cite{He01187}. 
The difference is that in our approach, this is ``scheduled'' in a total automated way by hyperprior learners. Another observation is that the highest learning rate is applied to BL-$1$. This actually encourages BL-$1$ to make an influence as significant as possible. It is very helpful to reduce meta gradient diminishing when unrolling and back-propagating gradients through many base-learning epochs (e.g., $100$ epochs in MTL).
Fig.\thinspace\ref{plot_mini_val_acc}(d) shows that 
BL-$1$ working at the initial epoch has the lowest values of $v$. In other words, BL-$1$ is almost disabled in the prediction of episode test.
Intriguingly, BL-$25$ instead of BL-$100$ gains the highest $v$ values. Our explanation is that during the base-learning, base-learners at latter epochs get more overfitted to the few training samples. Their functionality is thus suppressed. 
Note that our empirical results revealed that including the overfitted base-learners slightly improves the generalization capability of the approach.

%% file: tables/5_table_sota_new.tex
\begin{table*}[t]
\scriptsize
\begin{center}
\renewcommand\arraystretch{1.2}
\setlength{\tabcolsep}{1pt}{
\begin{tabular}{r|c|cc|ccc|ccc} 

\hline
\multirow{2.5}{*}{\textbf{Methods}} & \multirow{2.5}{*}{\textbf{Backbone}} & \multicolumn{2}{c}{\textbf{\emph{mini}ImageNet}} && \multicolumn{2}{c}{\textbf{\emph{tiered}ImageNet}} && \multicolumn{2}{c}{\textbf{FC100}}\\
\cmidrule{3-4}\cmidrule{6-7}\cmidrule{9-10}
&& \textbf{$1$-shot} & \textbf{$5$-shot} && \textbf{$1$-shot} & \textbf{$5$-shot} && \textbf{$1$-shot} & \textbf{$5$-shot}\\
\hline\hline
MatchNets~\cite{VinyalsBLKW16} & 4CONV & 43.44 & 55.31  && -- & -- && -- & -- \\
ProtoNets~\cite{SnellSZ17}  & 4CONV &  49.42 & 68.20 && 53.31 & 72.69 && -- & -- \\
MAML${}^{\diamond}$~\cite{FinnAL17} & 4CONV & 48.70 & 63.11 && 49.0  & 66.5 && 38.1  & 50.4 \\
MAML++${}^{\diamond}$~\cite{AntoniouICLR19} & 4CONV & 52.15 & 68.32  && 51.5 & 70.6  && 38.7  & 52.9\\
\hline
TADAM~\cite{OreshkinNIPS18} & ResNet-12 & 58.5  & 76.7  && -- & -- && 40.1 & 56.1 \\
MetaOptNet~\cite{lee2019meta} & ResNet-12 &  62.64 & 78.63 && 65.99 & 81.56 && 41.1 & 55.5  \\
CAN~\cite{hou2019cross} & ResNet-12 & {{63.85}} & 79.44 && \bluetf{\textbf{69.89}}  & {{84.23}} && -- & -- \\
CTM~\cite{li2019finding} & ResNet-18 & \bluetf{\textbf{64.12}} & 80.51 && {{68.41}}  & \bluetf{\textbf{84.28}} && -- & -- \\
MTL~\cite{SunCVPR2019} & ResNet-12 &  61.2 & 75.5 && -- & -- && \redtf{\textbf{45.1}} & 57.6 \\
MTL${}^{\diamond}$~\cite{SunCVPR2019} & ResNet-25 &  63.4 & 80.1 && {{69.1}} & 84.2 && 43.7 & \bluetf{\textbf{60.1}} \\
LEO~\cite{Rusu2019} & WRN-28-10 & 61.76 &77.59 && 66.33 & 81.44 && -- & -- \\
Robust20-dist${}^{\ddag}$~\cite{dvornik2019diversity}& WRN-28-10 &  63.28 &  \redtf{\textbf{81.17}} && -- & -- && -- & -- \\
\hline
\textbf{MAML+E$^3$BM} & 4CONV &  {{53.2}}($\uparrow${4.5}) & {{65.1}}($\uparrow${2.0}) && {{52.1}}($\uparrow${3.1}) & {{70.2}}($\uparrow${3.7}) && 39.9($\uparrow${1.8}) & {{52.6}}($\uparrow${2.2}) \\
(\textbf{+}time, \textbf{+}param)& -- & (8.9, 2.2) & (9.7, 2.2)&& (10.6, 2.2) & (9.3, 2.2) && (7.8, 2.2) & (12.1, 2.2)\\
\textbf{MTL+E$^3$BM} & ResNet-25 &  \redtf{\textbf{64.3}}($\uparrow${0.9}) & \bluetf{\textbf{81.0}}($\uparrow${0.9}) && \redtf{\textbf{70.0}}($\uparrow${0.9}) & \redtf{\textbf{85.0}}($\uparrow${0.8}) && \bluetf{\textbf{45.0}}($\uparrow${1.3}) & \redtf{\textbf{60.5}}($\uparrow${0.4}) \\
(\textbf{+}time, \textbf{+}param)& -- & (5.9, 0.7) & (10.2, 0.7)&& (6.7, 0.7) & (9.5, 0.7) && (5.7, 0.7) & (7.9, 0.7)\\
 \hline 
\multicolumn{10}{c}{\textbf{(a) Inductive Methods}}\\ 
\multicolumn{10}{c}{~}\\ 

 \hline 
EGNN~\cite{kim2019edge} & ResNet-12 & 64.02 & 77.20 && 65.45 & 82.52 && -- & -- \\
CAN+T~\cite{hou2019cross} & ResNet-12 & {{67.19}} & \bluetf{\textbf{80.64}} && {{73.21}}  & \redtf{\textbf{84.93}} && -- & -- \\
SIB${}^{\diamond\ddag}$~\cite{hu20empirical} & WRN-28-10 & \bluetf{\textbf{70.0}} & 79.2 && \bluetf{\textbf{72.9}} & 82.8 && \bluetf{\textbf{45.2}} &  \bluetf{\textbf{55.9}} \\
  \hline 
\textbf{SIB+E$^3$BM}${}^{\ddag}$& WRN-28-10 &  \redtf{\textbf{71.4}}($\uparrow${1.4}) & \redtf{\textbf{81.2}}($\uparrow${2.0}) && \redtf{\textbf{75.6}}($\uparrow${2.7})  & \bluetf{\textbf{84.3}}($\uparrow${1.5}) && \redtf{\textbf{46.0}}($\uparrow${0.8}) & \redtf{\textbf{57.1}}($\uparrow${1.2}) \\
(\textbf{+}time, \textbf{+}param)& --  & (2.1, 0.04) & (5.7, 0.04)&& (5.2, 0.04) & (4.9, 0.04) && (6.1, 0.04) & (7.3, 0.04)\\
 \hline 
 \multicolumn{10}{c}{\textbf{(b) Transductive Methods}}\\ 
 \multicolumn{10}{c}{~}\\ 
 \multicolumn{10}{l}{${}^{\diamond}$Our implementation on \emph{tiered}ImageNet and FC100.  \ \ ${}^{\ddag}$Input image size: $80\times80\times3$.}\\
\end{tabular}}
\end{center}
\caption{The $5$-class few-shot classification accuracies (\%) on \emph{mini}ImageNet, \emph{tiered}ImageNet, and FC100. ``(\textbf{+}time, \textbf{+}param)'' denote the additional computational time (\%) and parameter size (\%), respectively, when plugging-in E$^3$BM to baselines (MAML, MTL and SIB).
``--'' means no reported results in original papers. The \redtf{\textbf{best}} and \bluetf{\textbf{second best}} results are highlighted. 
} 
\vspace{-4mm}
\label{tab_e3bm_sota}
\end{table*}

%% file: tables/5_table_e3bm_ablation_new.tex
\begin{table*}[t]
\scriptsize
\begin{center}
\renewcommand\arraystretch{1.2}
\setlength{\tabcolsep}{2pt}{
\begin{tabular}{r|ccc|ccc|ccc|ccc} 
\hline
\multirow{2.5}{*}{\textbf{No.}} &\multicolumn{3}{c}{\textbf{Setting}} && \multicolumn{2}{c}{\textbf{\emph{mini}ImageNet}} && \multicolumn{2}{c}{\textbf{\emph{tiered}ImageNet}} && \multicolumn{2}{c}{\textbf{FC100}}\\
\cmidrule{2-4}\cmidrule{6-7}\cmidrule{9-10}\cmidrule{12-13}
&Method &Hyperprior &Learning && \textbf{$1$-shot} & \textbf{$5$-shot} && \textbf{$1$-shot} & \textbf{$5$-shot} && \textbf{$1$-shot} & \textbf{$5$-shot}\\
\hline\hline
1 & MTL~\cite{SunCVPR2019} & -- & Ind.  && 63.4 & 80.1 && 69.1 & 84.2 && 43.7 & 60.1\\
2 & MTL+E$^3$BM & FC & Ind.  && 64.3 & 80.9 && 69.8 & 84.6 && 44.8 & 60.5\\
3 & MTL+E$^3$BM & FC & Tra.  && 64.7 & 80.7 && 69.7 & 84.9 && 44.7 & \redtf{\textbf{60.6}} \\
4 & MTL+E$^3$BM & LSTM & Ind.  && 64.3 & {{81.0}} && 70.0 & \bluetf{\textbf{85.0}} && 45.0 & 60.4 \\
5 & MTL+E$^3$BM & LSTM & Tra.  && 64.5 & \bluetf{\textbf{81.1}} && 70.2 & \redtf{\textbf{85.3}} && 45.1 & \redtf{\textbf{60.6}}\\
\hline
6 & SIB~\cite{hu20empirical} & -- & Tra.  && 70.0 & 79.2 && 72.9 & 82.8 && 45.2 & 55.9\\
7 & SIB+E$^3$BM & FC & Tra.  && \bluetf{\textbf{71.3}} & {{81.0}} && \bluetf{\textbf{75.2}} & 83.8 && \bluetf{\textbf{45.8}} & 56.3 \\
8 & SIB+E$^3$BM & LSTM & Tra.  && \redtf{\textbf{71.4}} & \redtf{\textbf{81.2}} && \redtf{\textbf{75.6}} & 84.3 && \redtf{\textbf{46.0}} & 57.1\\
 \hline 
\end{tabular}}
\end{center}
\caption{The $5$-class few-shot classification accuracies (\%) of using different hyperprior learners, on the \emph{mini}ImageNet, \emph{tiered}ImageNet, and FC100. ``Ind.'' and ``Tra.'' denote the inductive and transductive settings, respectively. The \redtf{\textbf{best}} and \bluetf{\textbf{second best}} results are highlighted.}  
\vspace{-3mm}
\label{tab_e3bm_ablation}
\end{table*}

%% file: misc/5_mini_1shot_valacc.tex
\begin{figure}
\begin{center}
\subfigure[Val. acc. for $\alpha$]{\includegraphics[height=1.15in]{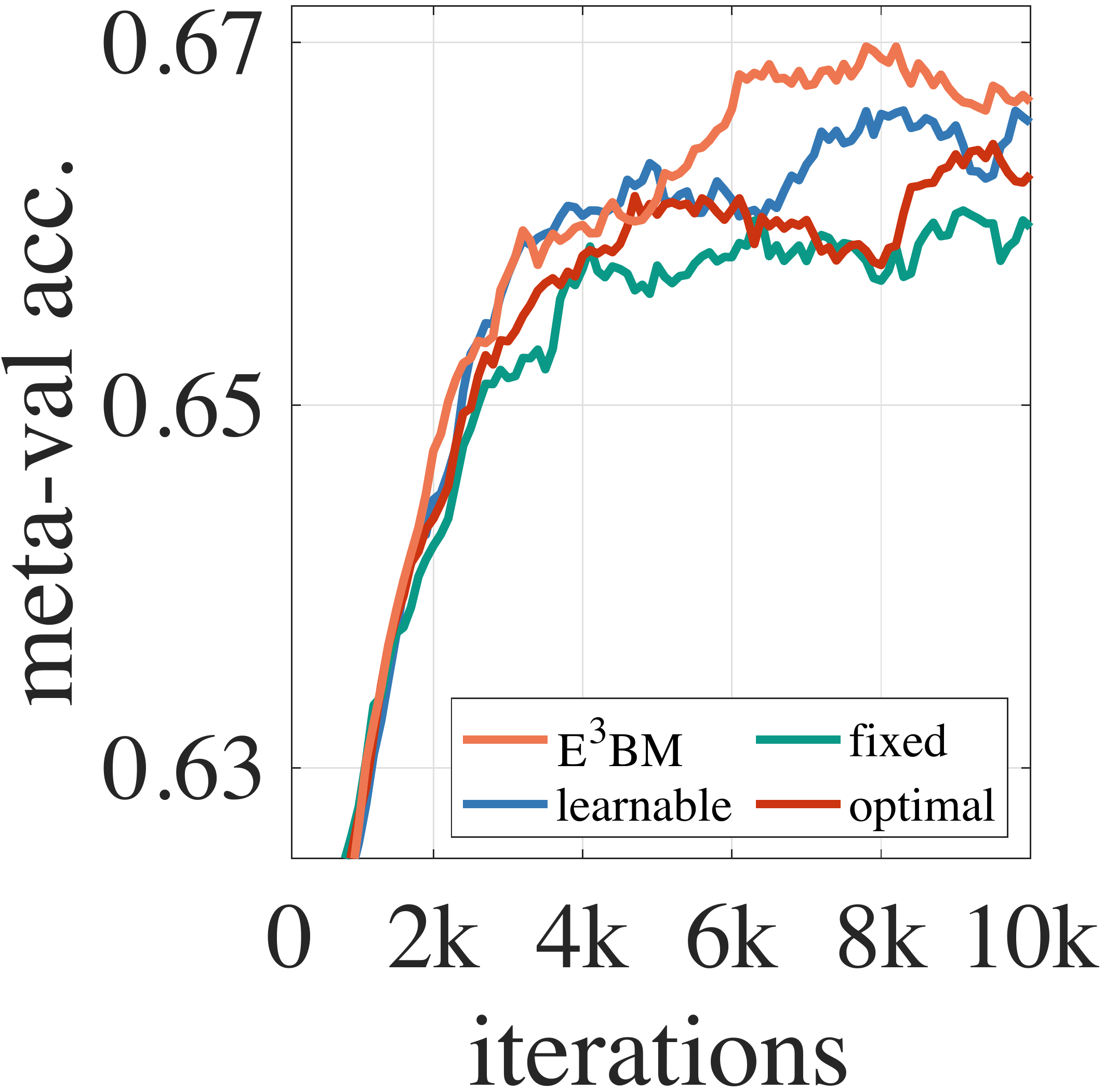}}
\subfigure[Val. acc. for $v$]{\includegraphics[height=1.15in]{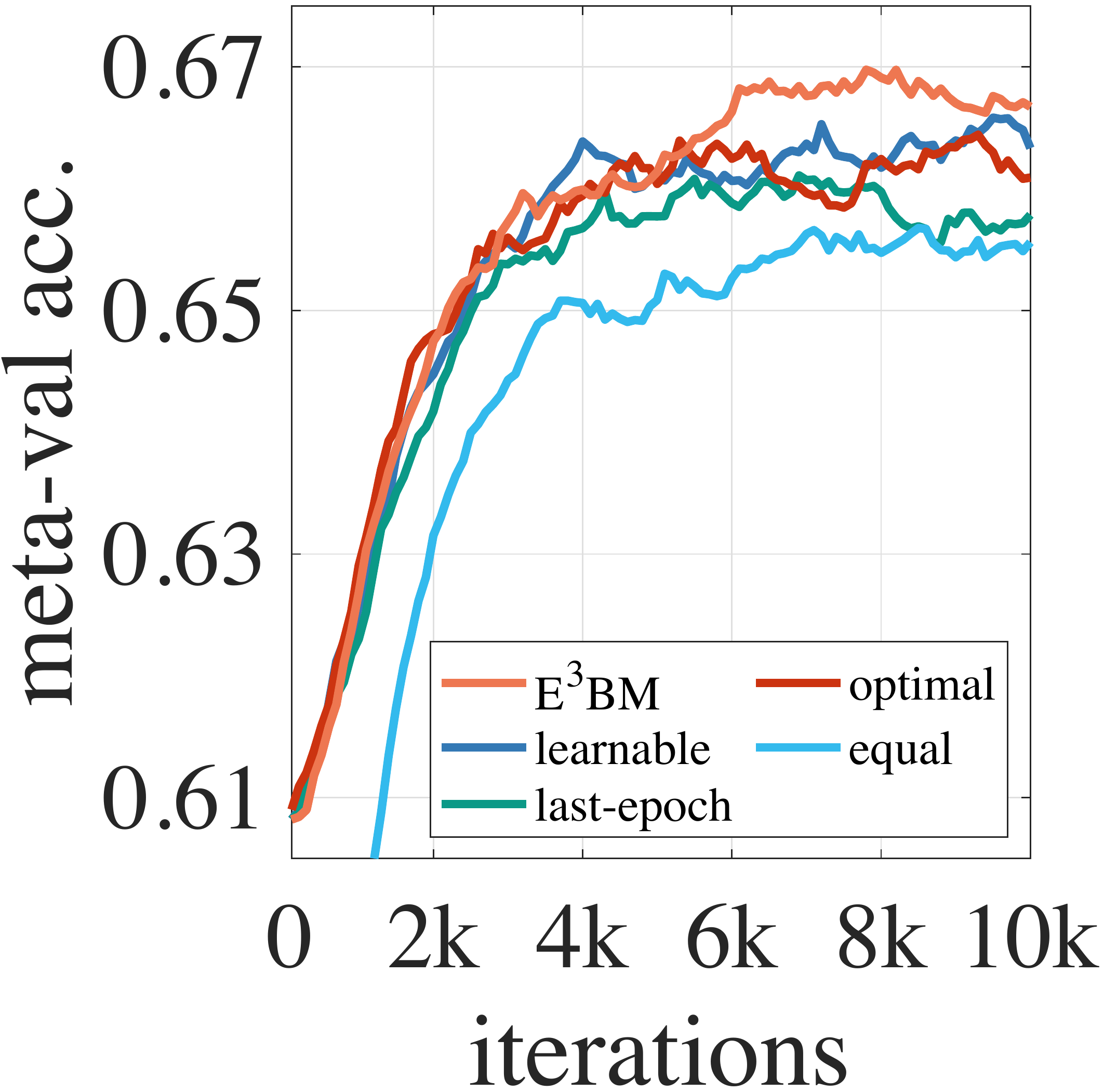}}
\subfigure[Values of $\alpha$]{\includegraphics[height=1.15in]{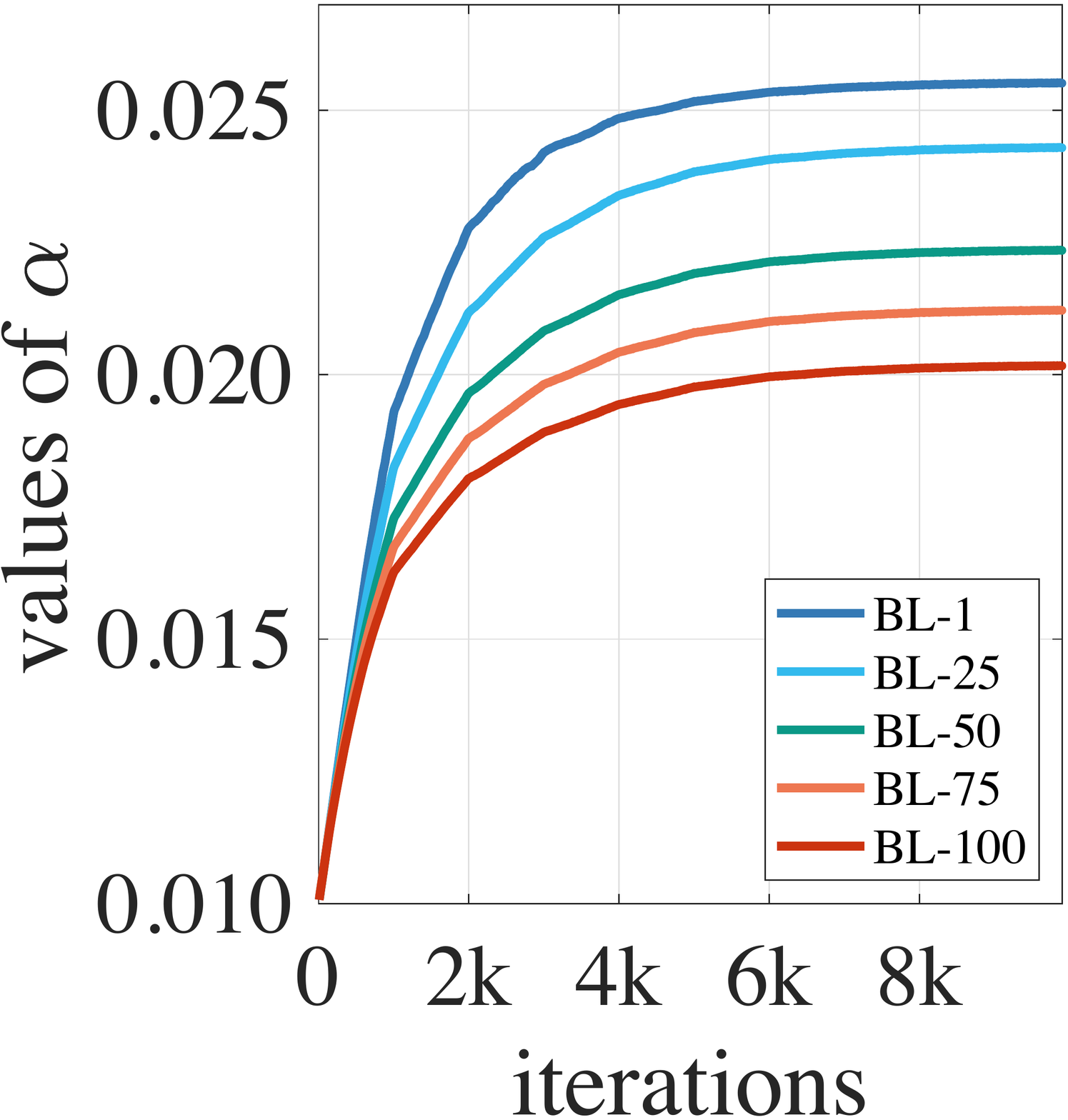}}
\subfigure[Values of $v$]{\includegraphics[height=1.15in]{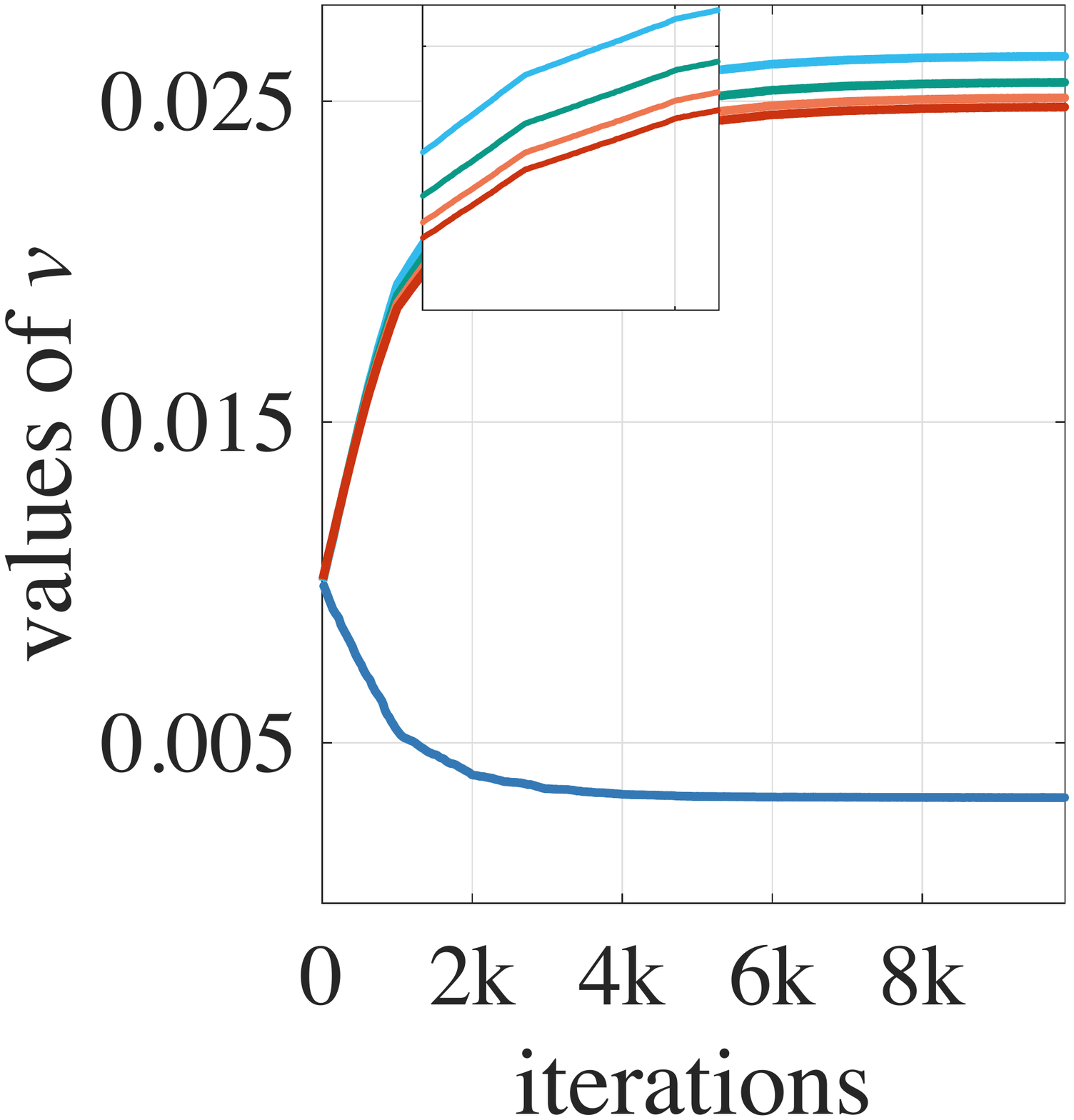}}
\end{center}
\caption{(a)(b): The meta-validation accuracies of ablation models.
The legends are explained in (a1)-(a4) and (v1)-(v5) in Sec.~\ref{sec_dataset} \textbf{Ablation settings}.
All curves are smoothed with a rate of $0.9$ for a better visualization. 
(c)(d): The values of $\alpha$ and $v$ generated by $\Psi_{\alpha}$ and $\Psi_v$, respectively. The setting is using MTL+E$^3$BM, ResNet-25, on \emph{mini}ImageNet, $1$-shot. 
}
\label{plot_mini_val_acc}
\end{figure}

%% file: sections/6_conclusion.tex
\section{Conclusions}

We propose a novel E$^3$BM approach that tackles the few-shot problem with an ensemble of epoch-wise base-learners that are trained and combined with task-specific hyperparameters.
In specific, E$^3$BM meta-learns the hyperprior learners to generate such hyperparameters conditional on the images as well as the training states for each episode.
Its resulting model allows to make use of multiple base-learners for more robust predictions.
It does not change the basic training paradigm of episodic few-shot learning, and is thus \emph{generic} and easy to plug-and-play with existing methods.
By applying E$^3$BM to multiple baseline methods, e.g., MAML, MTL and SIB, we achieved top performance on three challenging few-shot image classification benchmarks, with little computation or parametrization overhead.

%% file: sections/7_supplementary.tex
\clearpage

\beginsupp
\setcounter{section}{0}
\setcounter{page}{1}
\renewcommand\thesection{\Alph{section}}
\noindent
{\Large {\textbf{Supplementary Materials}}}
\\

These supplementary materials include E$^3$BM algorithms, results with confidence intervals, the supplementary plots to Fig.\thinspace\ref{plot_mini_val_acc}, backbone architectures, implementation details, ablation results for MAML, the inference time and the number of parameters, and the execution steps of our source code with PyTorch.

\section{E$^3$BM algorithms}
\label{sec_alg}
Algorithm~\ref{alg_AT} summarizes the meta-training (line 1-10) and meta-testing (line 11-16) procedures in our E$^3$BM approach.
For clarity, the base-learning steps within a single episode are moved to Algorithm~\ref{alg_base}.
\input{algorithm/pseudocode1}
\input{algorithm/pseuducode2}

\section{Results with confidence intervals}
In Table\thinspace\ref{tab_e3bm_sota_supp}, we supplement the few-shot classification accuracy (\%) on \emph{mini}ImageNet, \emph{tiered}ImageNet, and FC100 ($5$-class) with confidence intervals.
\input{tables/7_table_all_results}

\section{Supplementary figures}
\myparagraph{Supplementary to Fig.\thinspace\ref{plot_mini_val_acc}(a)(b).}
In Fig.~\ref{plot_mini_5shot_val_acc}, we supplement the meta-validation accuracies for the $1$-shot and $5$-shot cases on \emph{mini}ImageNet, \emph{tiered}ImageNet, and FC100 (Note that Fig.\thinspace\ref{plot_mini_val_acc} already has the \emph{mini}ImageNet $1$-shot results).

\myparagraph{Supplementary to Fig.\thinspace\ref{plot_mini_val_acc}(c)(d).}
On the \emph{mini}ImageNet, we supplement the plots of $\alpha$ and $v$ in the $5$-shot case in Fig.\thinspace\ref{plot_all_alpha_v}(a)(b).
On the \emph{tiered}ImageNet, we show the plots of $\alpha$ and $v$ in Fig.\thinspace\ref{plot_all_alpha_v}(c)(d) and (e)(f), respectively for $1$-shot and $5$-shot cases. 
On the FC100, we show the plots of $\alpha$ and $v$ in Fig.\thinspace\ref{plot_all_alpha_v}(g)(h) and (i)(j), respectively for $1$-shot and $5$-shot cases.
Each figure demonstrates the values of $\alpha$ (or $v$) generated by the model ``MTL+E$^3$BM'' as in Table 1. 

\section{Backbone architectures}
\myparagraph{4CONV} consists of $4$ layers with $3\times 3$ convolutions and $32$ filters, followed by batch normalization (BN), a ReLU nonlinearity, and $2\times 2$ max-pooling.

\myparagraph{ResNet-12} has
$3$ residual blocks. 
Each block has $4$ convolution layers with $3\times 3$ kernels. The number of filters starts from $160$ and is doubled every next block. After a global average pooling layer, it gets a $640$-dim embedding. This architecture follows~\cite{lee2019meta}.

\myparagraph{ResNet-25} has
$3$ residual blocks after an initial convolution layer. 
Each block has $8$ convolution layers with $3\times 3$ kernels. The number of filters starts from $160$ and is doubled every next block. After a global average pooling layer, it gets a $640$-dim embedding. This architecture follows~\cite{ye2018learning}.

\myparagraph{WRN-28-10}
has its depth and width set to $28$ and $10$, respectively. %
After a global average pooling in the last layer of the backbone, it gets a $640$-dimensional embedding. 
For this backbone, we resize the input image to $80\times80\times3$ for a fair comparison with related methods~\cite{SunCVPR2019,hu20empirical}.  Other details are the same as those with ResNet-25~\cite{ye2018learning,Rusu2019}.

\section{Implementation details}

\myparagraph{MTL+E$^3$BM.} The meta learning rates for the scaling and shifting weights $\Phi_{SS}$ and the base-learner initializer $\theta$ are set to $1\times10^{-4}$ uniformly. The base learning rates $\{\alpha_m'\}_{m=1}^M$ (Fig.\thinspace\ref{figure_netarch_hyper}) are initialized as $1\times10^{-2}$~\cite{SunCVPR2019,ye2018learning}. We meta-train MTL+E$^3$BM for $10,000$ iterations and use the model, which has the highest meta-validation accuracy, for meta-test.

\myparagraph{SIB+E$^3$BM.} The meta learning rates for both SIB network $\phi(\lambda, \xi)$ and base-learner initializer $\theta$ are set to $1\times10^{-3}$ uniformly. The base learning rates $\{\alpha_m'\}_{m=1}^M$ (Fig.\thinspace\ref{figure_netarch_hyper}) are initialized as $1\times10^{-3}$~\cite{hu20empirical}.
We meta-train SIB+E$^3$BM for $50,000$ iterations and use the model, which has the highest meta-validation accuracy, for meta-test.

\myparagraph{MAML+E$^3$BM.} MAML only contains a model initializer $\theta$, and we set its meta-learning rate as $1\times10^{-3}$~\cite{FinnAL17}.  
The base learning rates $\{\alpha_m'\}_{m=1}^M$ (Fig.\thinspace\ref{figure_netarch_hyper}) are initialized as $1\times10^{-3}$. We meta-train MAML+E$^3$BM for $60,000$ iterations and use the model, which has the highest meta-validation accuracy, for meta-test.

\myparagraph{Shared hyperparameters.}
The meta learning rates for $\Psi_{\alpha}$ and $\Psi_v$ are set to $1\times10^{-6}$ uniformly.
For initializing $\{v_m'\}_{m=1}^M$ (Fig.\thinspace\ref{figure_netarch_hyper}), we have two options. One is each $v_m'$ is initialized as $1$/(number of base-learners), and the other one is that $\{v_m'\}_{m=1}^{M-1}$ are initialized as $0$ and $v_M'$ as $1$. 
In Eq.~\eqref{eq_psi_final} in Sec.~\ref{subsec_hyperprior}, $\lambda_1$ and $\lambda_2$ are set to $1\times10^{-4}$. For the rest of the hyperparameters, we follow the original settings of baselines~\cite{FinnAL17,hu20empirical,SunCVPR2019}.

\myparagraph{Constraints for $v$ and $\alpha$.}
In the constraint mode, we applied the constraints on $v$ and $\alpha$ to force them to be positive and smaller than $1$. We did not have any constraint for $\Delta v$ or $\Delta \alpha$. Please note that the constraints are not applied in the default setting.

\myparagraph{Dataloader} For MAML, we use the same dataloader as~\cite{FinnAL17}. For MTL, we follow~\cite{ye2018learning,Zhang_2020_CVPR,DeepEMDJ}. For SIB, we follow~\cite{hu20empirical}.

\section{Ablation results for MAML}
In Table\thinspace\ref{tab_e3bm_ablation_supp}, we supplement the ablation results for ``MAML+E$^3$BM'' on \emph{mini}ImageNet, \emph{tiered}ImageNet, and FC100 ($5$-class).
\input{tables/7_table_e3bm_ablation_supp}

\section{The inference time and the number of parameters}
In Table\thinspace\ref{tab_e3bm_time_supp}, we supplement the the inference time and the number of parameters of baselines ($100$ epochs, \emph{mini}ImageNet, $5$-way $1$-shot, on NVIDIA V100 GPU)
\input{tables/7_table_time}

\section{Executing the source code with PyTorch}

We provide our PyTorch code at \href{https://gitlab.mpi-klsb.mpg.de/yaoyaoliu/e3bm}{https://gitlab.mpi-klsb.mpg.de/yaoyaoliu/e3bm}. 
To run this repository, we kindly advise you to install python 3.6 and PyTorch 1.2.0 with Anaconda.
You may download Anaconda and read the installation instruction on the official website (\href{https://www.anaconda.com/download/}{https://www.anaconda.com/download/}).

\noindent Create a new environment and install PyTorch and torchvision on it:
\begin{lstlisting}[frame=shadowbox]
conda create --name e3bm-pytorch python=3.6
conda activate e3bm-pytorch
conda install pytorch=1.2.0 
conda install torchvision -c pytorch
\end{lstlisting}

\noindent Install other requirements:
\begin{lstlisting}[frame=shadowbox]
pip install -r requirements.txt
\end{lstlisting}

\noindent Run meta-training with default settings (data and pre-trained model will be downloaded automatically):
\begin{lstlisting}[frame=shadowbox]
python main.py -backbone resnet12 -shot 1 -way 5 -mode meta_train -dataset miniimagenet
python main.py -backbone resnet12 -shot 5 -way 5 -mode meta_train -dataset miniimagenet
python main.py -backbone resnet12 -shot 1 -way 5 -mode meta_train -dataset tieredimagenet
python main.py -backbone resnet12 -shot 5 -way 5 -mode meta_train -dataset tieredimagenet
\end{lstlisting}

\noindent Run pre-training with default settings:
\begin{lstlisting}[frame=shadowbox]
python main.py -backbone resnet12 -mode pre_train -dataset miniimagenet
python main.py -backbone resnet12 -mode pre_train -dataset tieredimagenet
\end{lstlisting}


\input{misc/5_all_valacc}
\input{misc/5_all_value_alpha_v}

%% file: algorithm/pseudocode1.tex
\begin{algorithm}
\caption{An Ensemble of Epoch-wise Empirical Bayes Models (E$^3$BM)}
\label{alg_AT}
\SetAlgoLined
\SetKwInput{KwData}{\% Meta-train phase}
\SetKwInput{KwResult}{\% Meta-test phase}
 \KwIn{Meta-train episode distribution $p_{tr}(\mathcal{T})$, Meta-test episode distribution $p_{te}(\mathcal{T})$, and meta-train stepsizes $\beta_1$ and $\beta_2$.}
 \KwOut{The average accuracy of meta-test.}
 \KwData{}
 Randomly initialize $\theta$\;
 \For{all meta iterations}{
 Sample a batch of meta-train episodes $\{\mathcal{T}_i\} \in p_{tr}(\mathcal{T})$\;
 \For{$\mathcal{T}_i$ in $\{\mathcal{T}_i\}$}{
 Train the sequence of base-learners on $\mathcal{T}_i$ by \textbf{Algorithm}~\ref{alg_base}\;
 }
 Evaluate $\mathcal{L}^{(te)}$ with Eq.~\eqref{eq_epite_loss_weighted} \;
 Optimize $\Psi_{\alpha}$, and $\Psi_v$ with Eq.~\eqref{eq_update_alpha} using $\beta_1$ and $\beta_2$\;
 Optimize other meta components, e.g., $\theta$.
 }
 \KwResult{}
 Sample meta-test episodes $\{\mathcal{T}_i\} \in p_{te}(\mathcal{T})$\;
 \For{$\mathcal{T}_i$ in $\{\mathcal{T}_i\}$}{
 Train the sequence of base-learners on $\mathcal{T}_i$ and obtain the prediction scores $\hat{y}^{(te)}$ by \textbf{Algorithm}~\ref{alg_base}\;
 Compute episode test accuracy $Acc_i$\;
 }
 Return the average accuracy of $\{Acc_i\}$.
\end{algorithm}

%% file: algorithm/pseuducode2.tex
\begin{algorithm}
\caption{Learning the ensemble of base-learners in one episode}
\label{alg_base}
\SetAlgoLined
\SetKwInput{KwData}{Input}
\SetKwInput{KwResult}{Output}
 \KwData{An episode $\mathcal{T}$, hyperprior learners $\Psi_{\alpha}$ and $\Psi_v$.}
 \KwResult{Prediction $\hat{y}^{(te)}$, and episode test loss $\mathcal{L}^{(te)}$.}
 Initialize $\Theta_0=\theta$\;
 \For{$m$ in $\{1, ..., M\}$}{
 Evaluate $\mathcal{L}^{(tr)}_m$ with Eq.~\eqref{eq_base_loss_normal} and compute $\nabla_{\Theta}\mathcal{L}^{(tr)}_m$\;
 Get $\alpha_m$ from $\Psi_{\alpha}$ and get $v_m$ from $\Psi_v$\;
 Get $\Theta_m$ using $\alpha_m$ with Eq.~\eqref{eq_theta_normal}\;
 Compute $z_m$ with Eq.~\eqref{MAML_pred}\;
 \eIf{$m=1$}{
 Initialize $\hat{y}^{(te)}_1=v_1 z_1$\;
 }
 {Compute $\hat{y}^{(te)}_m$ using $v_m$ with Eq.~\eqref{yhat_update}\;
 }
 }
 Evaluate $\mathcal{L}^{(te)}$ with Eq.~\eqref{eq_epite_loss_weighted}.
\end{algorithm}

%% file: tables/7_table_all_results.tex
\begin{table*}[t]
\scriptsize
\begin{center}
\renewcommand\arraystretch{1.2}
\setlength{\tabcolsep}{1pt}{
\begin{tabular}{r|c|cc|ccc|ccc} 

\hline
\multirow{2.5}{*}{\textbf{Methods}} & \multirow{2.5}{*}{\textbf{Backbone}} & \multicolumn{2}{c}{\textbf{\emph{mini}ImageNet}} && \multicolumn{2}{c}{\textbf{\emph{tiered}ImageNet}} && \multicolumn{2}{c}{\textbf{FC100}}\\
\cmidrule{3-4}\cmidrule{6-7}\cmidrule{9-10}
&& \textbf{$1$-shot} & \textbf{$5$-shot} && \textbf{$1$-shot} & \textbf{$5$-shot} && \textbf{$1$-shot} & \textbf{$5$-shot}\\
\hline\hline

\textbf{MAML+E$^3$BM} & 4CONV &  {{53.2}} \tiny{$\pm$ 1.8} & {{65.1}} \tiny{$\pm$ 0.9} && {{52.1}} \tiny{$\pm$ 1.8} & {{70.2}} \tiny{$\pm$ 0.9} && 39.9 \tiny{$\pm$ 1.8} & {{52.6}} \tiny{$\pm$ 0.9} \\
\textbf{MTL+E$^3$BM} & ResNet-12 &  {{63.8}} \tiny{$\pm$ 0.4} & {{80.1}} \tiny{$\pm$ 0.3} && {{71.2}} \tiny{$\pm$ 0.4} & {{85.3}} \tiny{$\pm$ 0.3} && {{43.2}} \tiny{$\pm$ 0.3} & {{60.2}} \tiny{$\pm$ 0.3} \\
\textbf{MTL+E$^3$BM} & ResNet-25 &  {{64.3}} \tiny{$\pm$ 0.4} & {{81.0}} \tiny{$\pm$ 0.3} && {{70.0}} \tiny{$\pm$ 0.4} & {{85.0}} \tiny{$\pm$ 0.3} && {{45.0}} \tiny{$\pm$ 0.4} & {{60.5}} \tiny{$\pm$ 0.3} \\
\textbf{SIB+E$^3$BM}& WRN-28-10 & {{71.4}} \tiny{$\pm$ 0.5} & {{81.2}} \tiny{$\pm$ 0.4} && {{75.6}} \tiny{$\pm$ 0.6}  & {{84.3}} \tiny{$\pm$ 0.4} && {{46.0}} \tiny{$\pm$ 0.6} & {{57.1}} \tiny{$\pm$ 0.4} \\
 \hline 

\end{tabular}}

\end{center}

\cotronlvsapcetablecaption
\caption{{\mycaptionsupp{Supplementary to Table\thinspace\ref{tab_e3bm_sota}.}} Few-shot classification accuracy (\%) on \emph{mini}ImageNet, \emph{tiered}ImageNet, and FC100 ($5$-class).
} 
\label{tab_e3bm_sota_supp}
\end{table*}

%% file: tables/7_table_e3bm_ablation_supp.tex
\begin{table*}[h]
\scriptsize
\begin{center}
\renewcommand\arraystretch{1.2}
\setlength{\tabcolsep}{2pt}{
\begin{tabular}{r|ccc|ccc|ccc|ccc} 

\hline 
\multirow{2.5}{*}{\textbf{No.}} &\multicolumn{3}{c}{\textbf{Setting}} && \multicolumn{2}{c}{\textbf{\emph{mini}ImageNet}} && \multicolumn{2}{c}{\textbf{\emph{tiered}ImageNet}} && \multicolumn{2}{c}{\textbf{FC100}}\\
\cmidrule{2-4}\cmidrule{6-7}\cmidrule{9-10}\cmidrule{12-13}
&Method &Hyperprior &Learning && \textbf{$1$-shot} & \textbf{$5$-shot} && \textbf{$1$-shot} & \textbf{$5$-shot} && \textbf{$1$-shot} & \textbf{$5$-shot}\\
\hline\hline
1 & MAML~\cite{FinnAL17} & -- & Ind.  && 48.70 & 63.11 && 49.0 & 66.5 && 38.1 & 50.4\\
2 & 5x MAML & -- & Ind.  && 52.1 & 65.1 && 51.1 & 68.8 && 40.1 & 50.8\\
3 & MAML+E$^3$BM & FC & Ind.  && 52.1 & 65.1 && 51.1 & 68.8 && 39.5 & 51.7\\
4 & MAML+E$^3$BM & FC & Tra.  && 52.8 & 65.3 && 52.2 & 69.5 && 40.4 & 51.8\\
5 & MAML+E$^3$BM & LSTM & Ind.  && 53.2 & 65.0 && 52.1 & 70.2 && 39.9 & 52.6\\
6 & MAML+E$^3$BM & LSTM & Tra.  && 53.8 & 65.2 && 52.7 & 70.5 && 40.4 & 52.3\\
 \hline 
\end{tabular}}
\end{center}
\caption{{\mycaptionsupp{Supplementary to Table~\ref{tab_e3bm_ablation}.}} Results (\%) for different hyperprior learners on \emph{mini}ImageNet, \emph{tiered}ImageNet, and FC100 ($5$-class). ``Ind.'' and ``Tra.'' denote inductive and transductive settings, respectively. 
}
\label{tab_e3bm_ablation_supp}
\end{table*}

%% file: tables/7_table_time.tex
\begin{table*}[h]
\scriptsize
\begin{center}
\renewcommand\arraystretch{1.2}
\setlength{\tabcolsep}{2pt}{
\begin{tabular}{l|cc|cc} 

\hline 
\textbf{No.} & \textbf{Method} & \textbf{Backbone} & \textbf{\# Param} & \textbf{Time (min)}\\
\hline\hline
1 & MTL & ResNet-25 & 4,321k & 73.3\\
2 & MTL+E$^3$BM (ours) & ResNet-25 & 4,351k & 77.6\\
3 & SIB & WRN-28-10 & 36,475k & 325.0\\
4 & SIB+E$^3$BM (ours) & WRN-28-10 & 36,490k & 331.8\\
5 & ProtoNets & ResNet-12 & 12,424k & 35.2\\
6 & MatchNets & ResNet-12 & 12,424k & 37.3\\
7 & ProtoNets & ResNet-25 & 36,579k & 70.5\\
7 & ProtoNets & WRN-28-10 & 36,482k & 350.1\\

 \hline 
\end{tabular}}
\end{center}

\caption{{\mycaptionsupp{Supplementary to Table~\ref{tab_e3bm_sota}.}} The inference time and the number of parameters of baselines ($100$ epochs, \emph{mini}ImageNet, $5$-way $1$-shot, on NVIDIA V100 GPU). 
}
\label{tab_e3bm_time_supp}
\end{table*}

%% file: misc/5_all_valacc.tex
\begin{figure}
\begin{center}
\subfigure[$v$, \emph{mini}ImageNet, $5$-shot]{\includegraphics[height=1.15in]{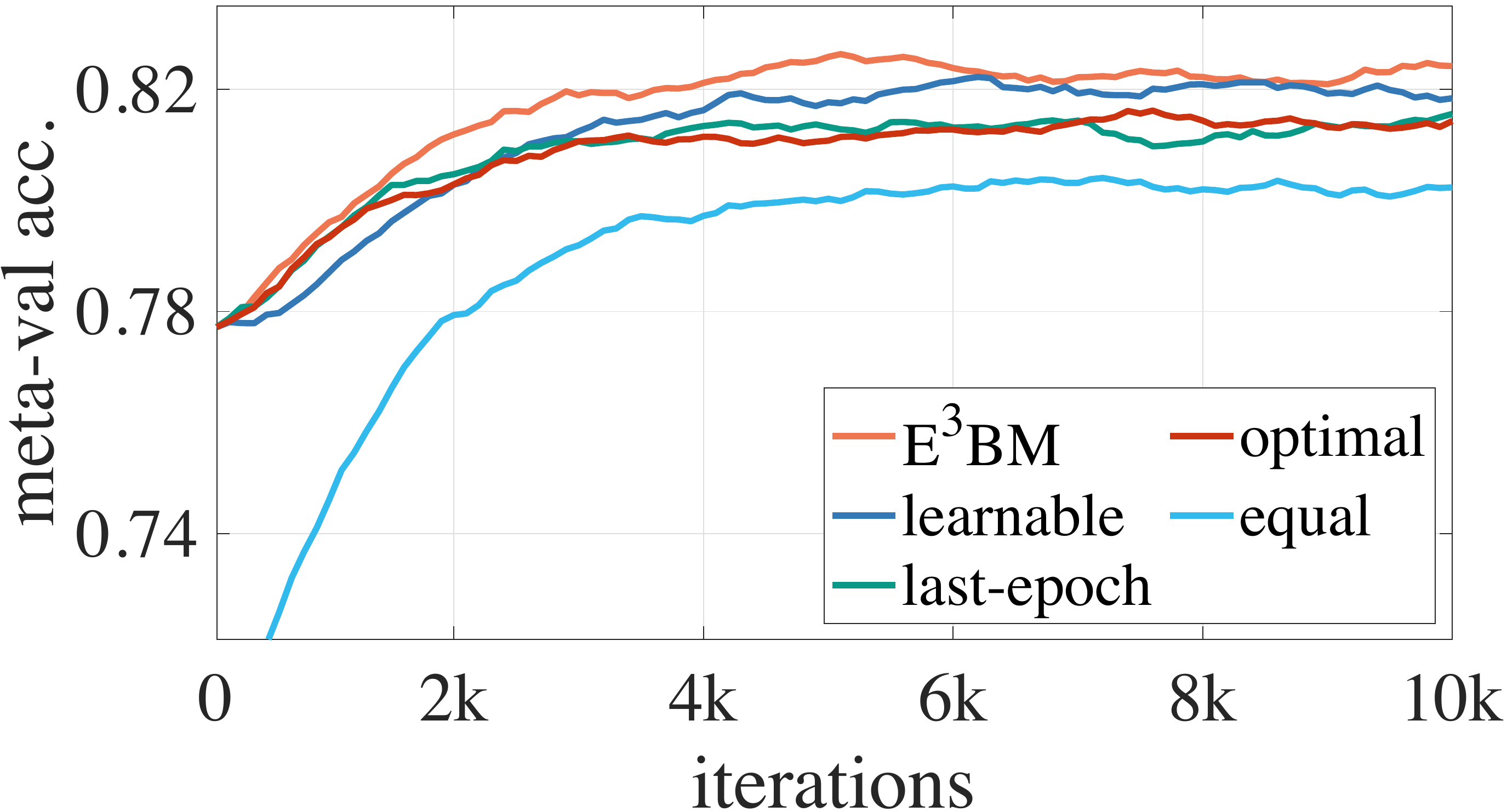}}
\ \ \ \ \ \ \ \ 
\subfigure[$\alpha$, \emph{mini}ImageNet, $5$-shot]{\includegraphics[height=1.15in]{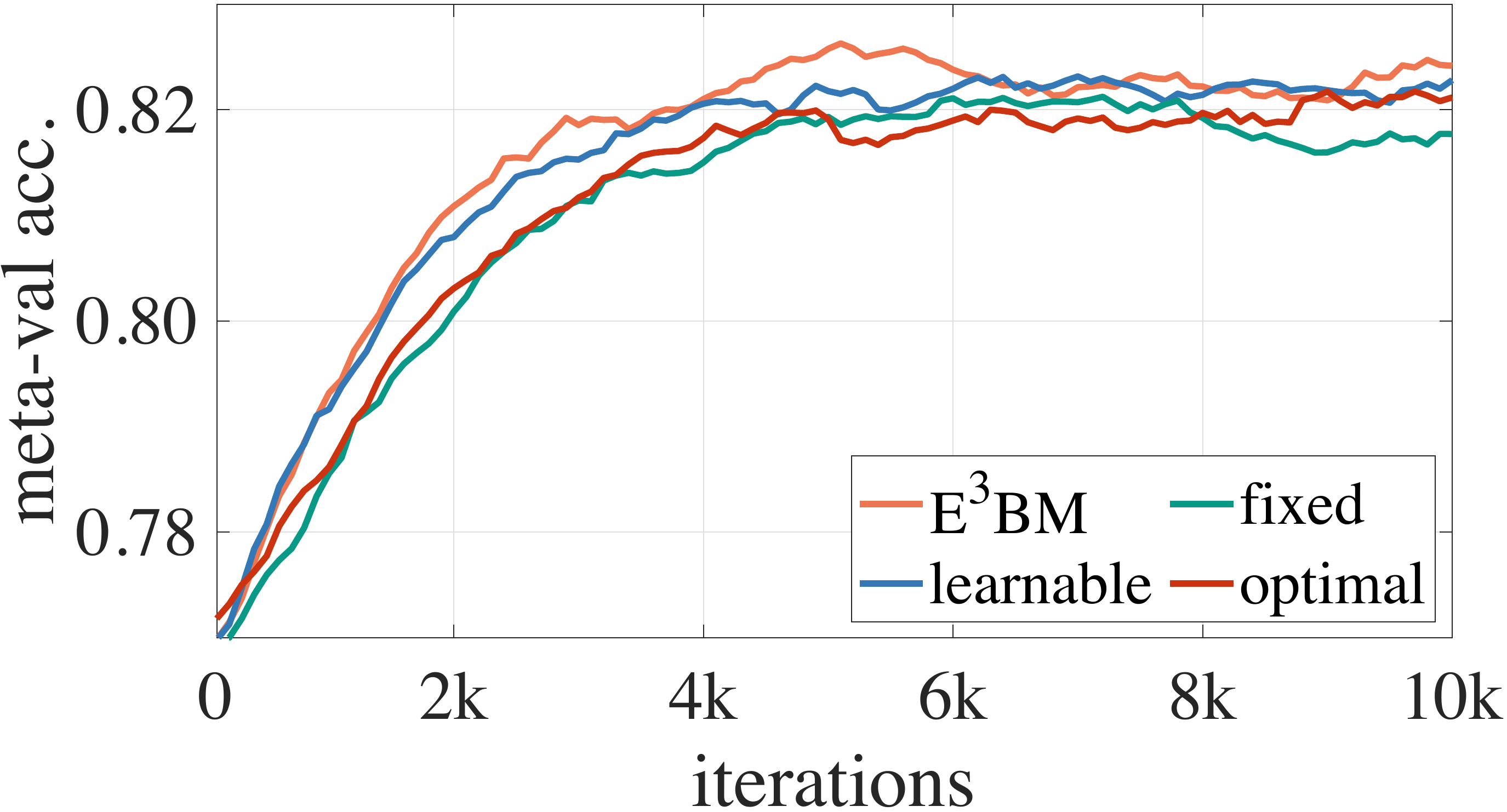}}

\subfigure[$v$, \emph{tiered}ImageNet, $1$-shot]{\includegraphics[height=1.15in]{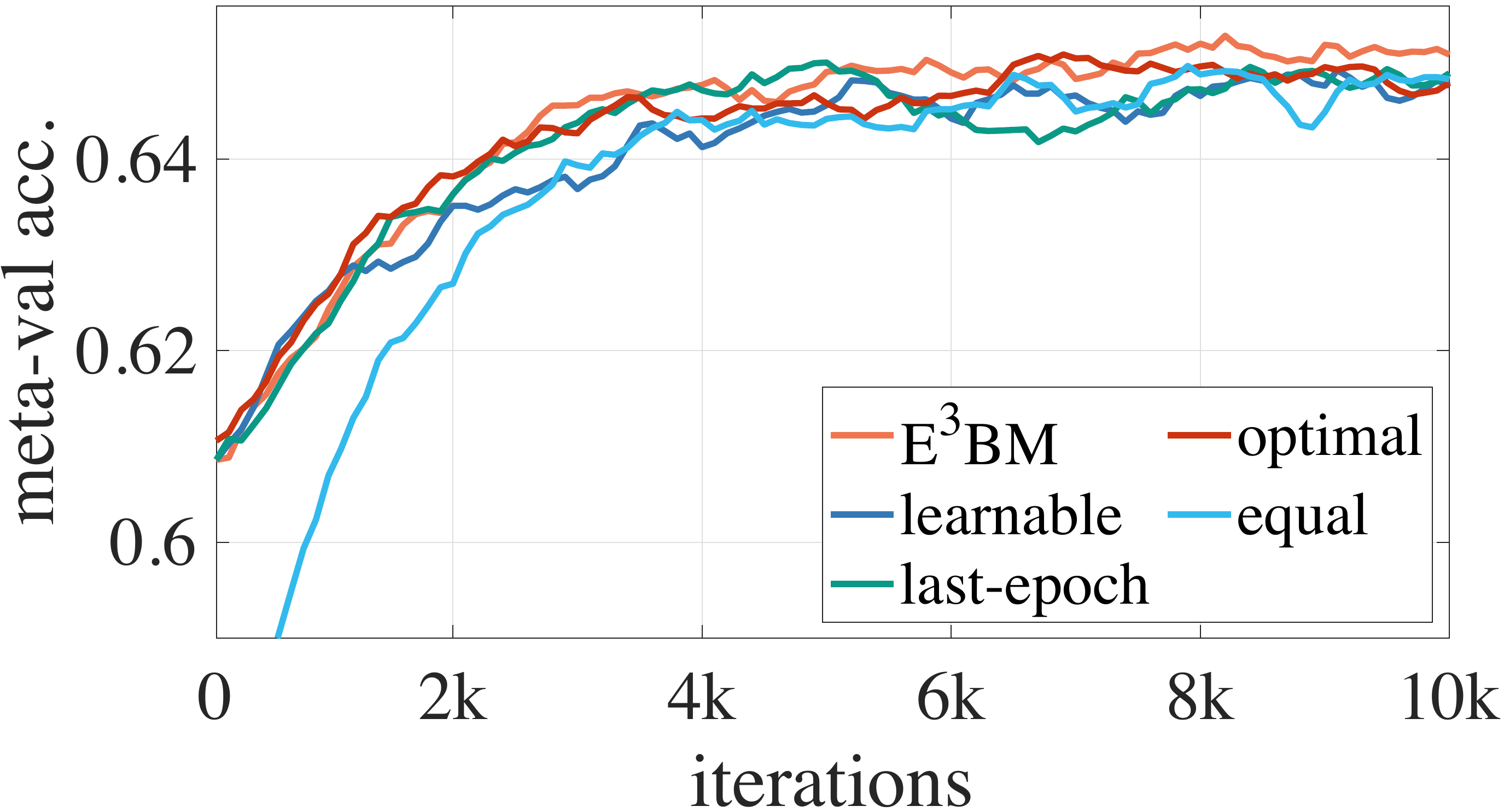}}
\ \ \ \ \ \ \ \ 
\subfigure[$\alpha$, \emph{tiered}ImageNet, $1$-shot]{\includegraphics[height=1.15in]{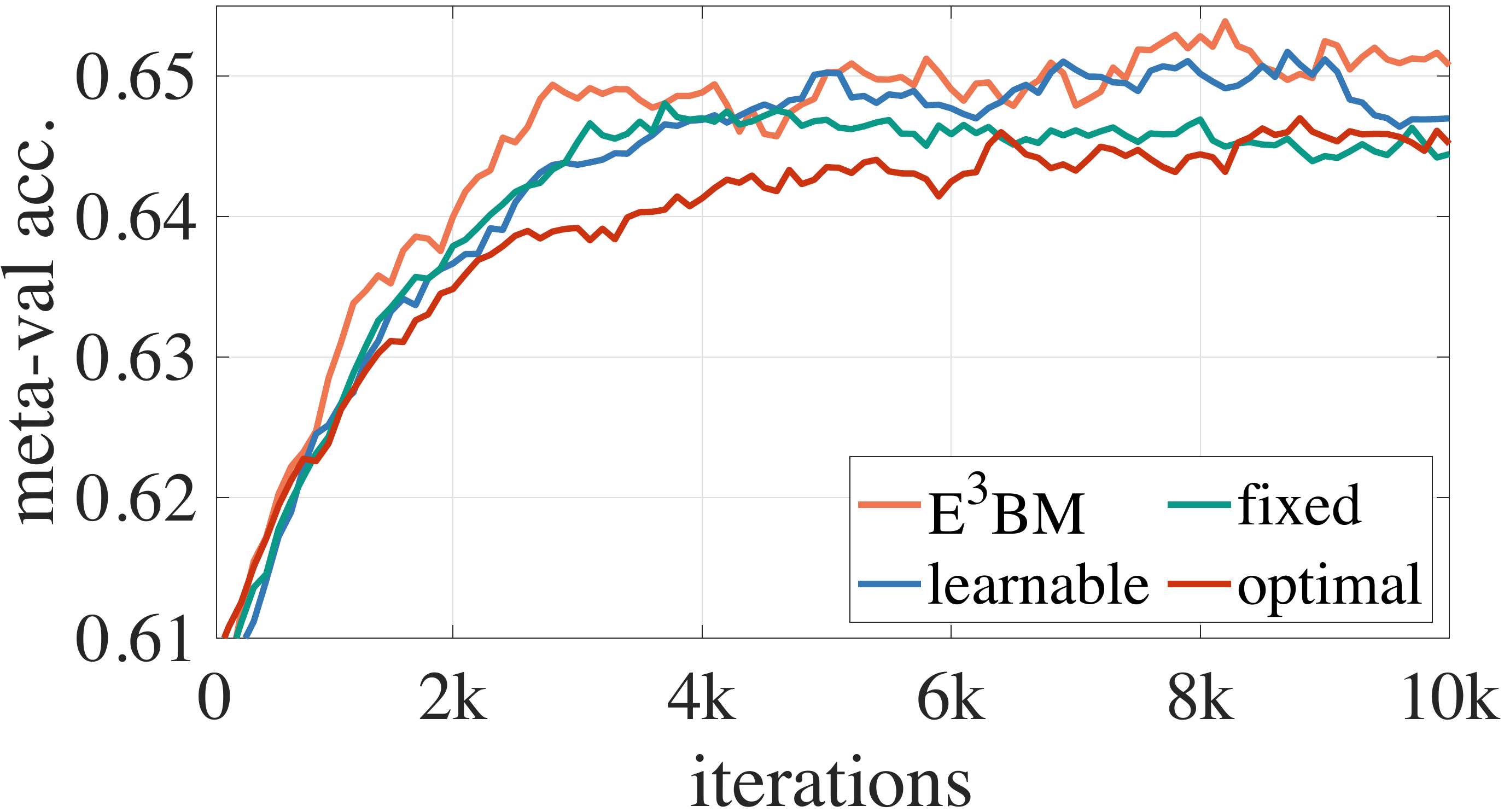}}

\subfigure[$v$, \emph{tiered}ImageNet, $5$-shot]{\includegraphics[height=1.15in]{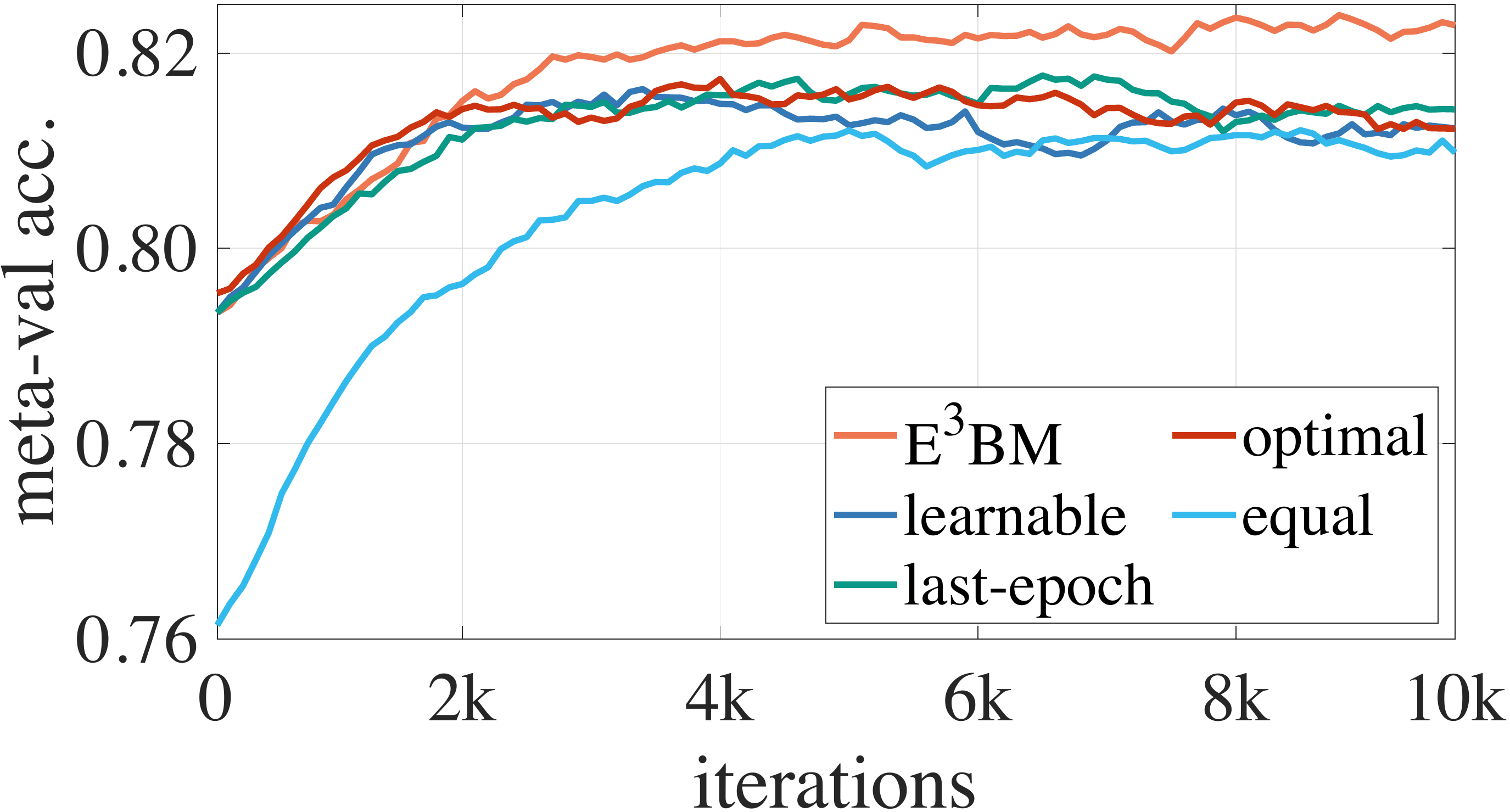}}
\ \ \ \ \ \ \ \ 
\subfigure[$\alpha$, \emph{tiered}ImageNet, $5$-shot]{\includegraphics[height=1.15in]{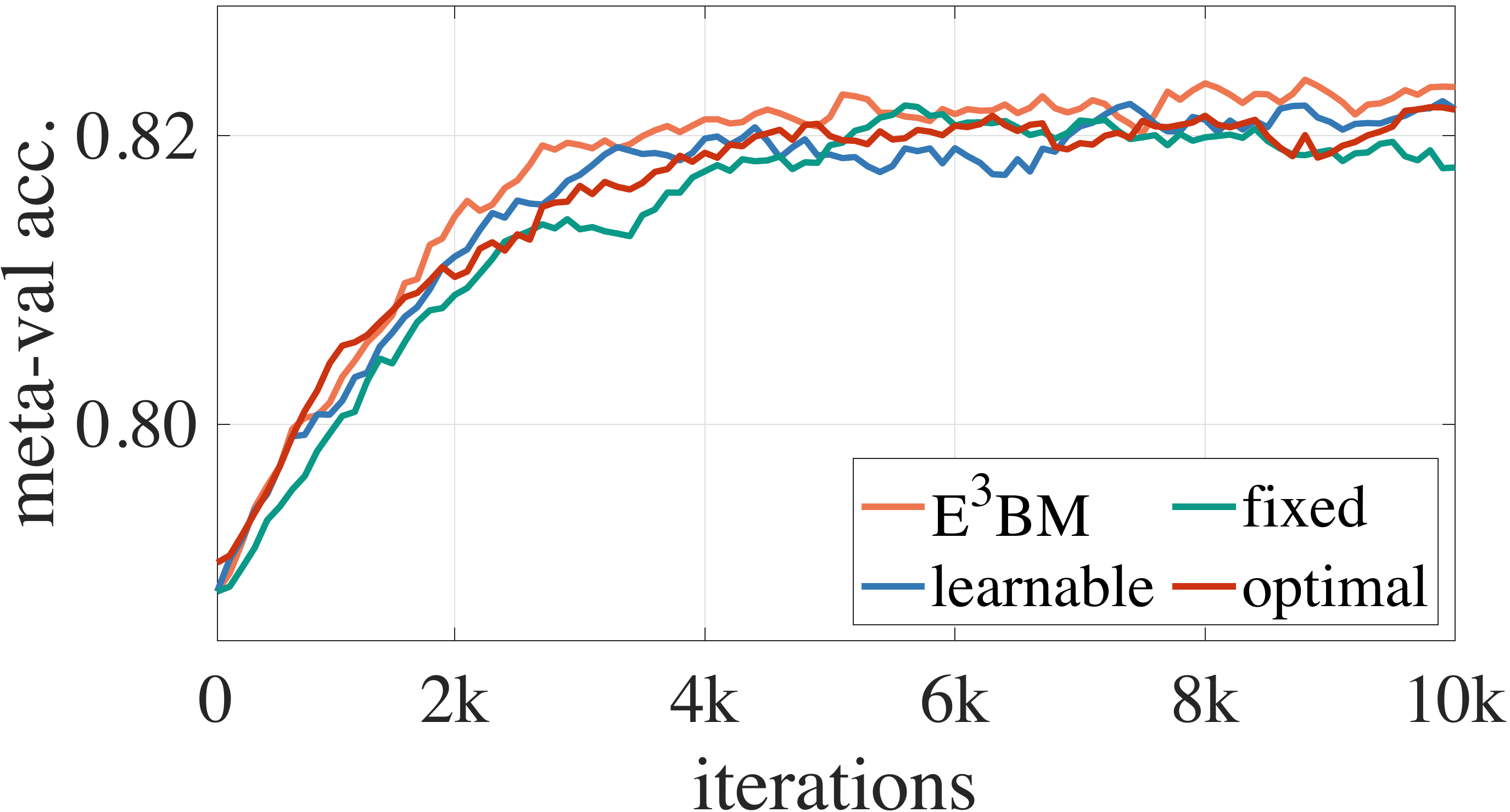}}

\subfigure[$v$, FC100, $1$-shot]{\includegraphics[height=1.15in]{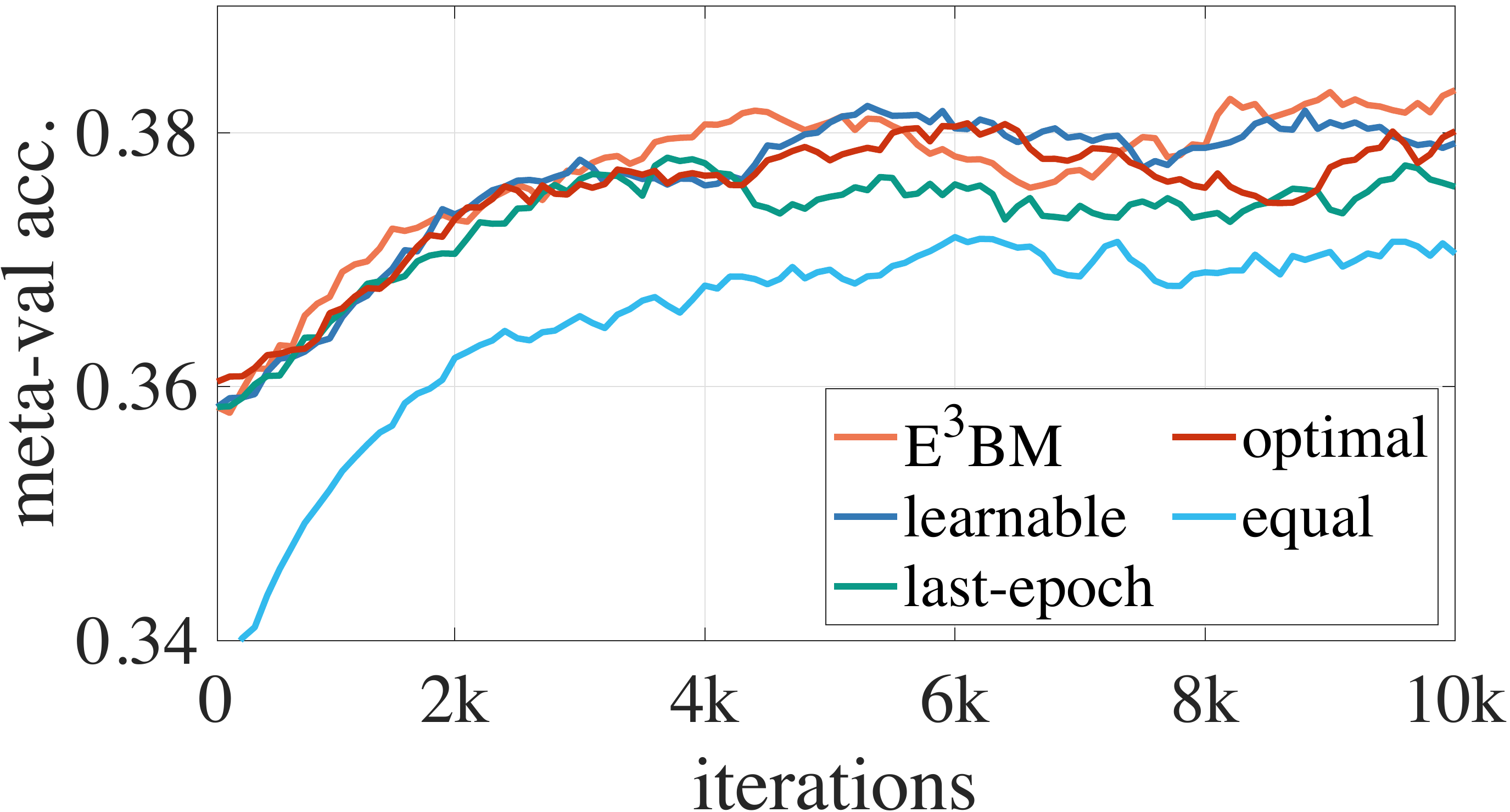}}
\ \ \ \ \ \ \ \ 
\subfigure[$\alpha$, FC100, $1$-shot]{\includegraphics[height=1.15in]{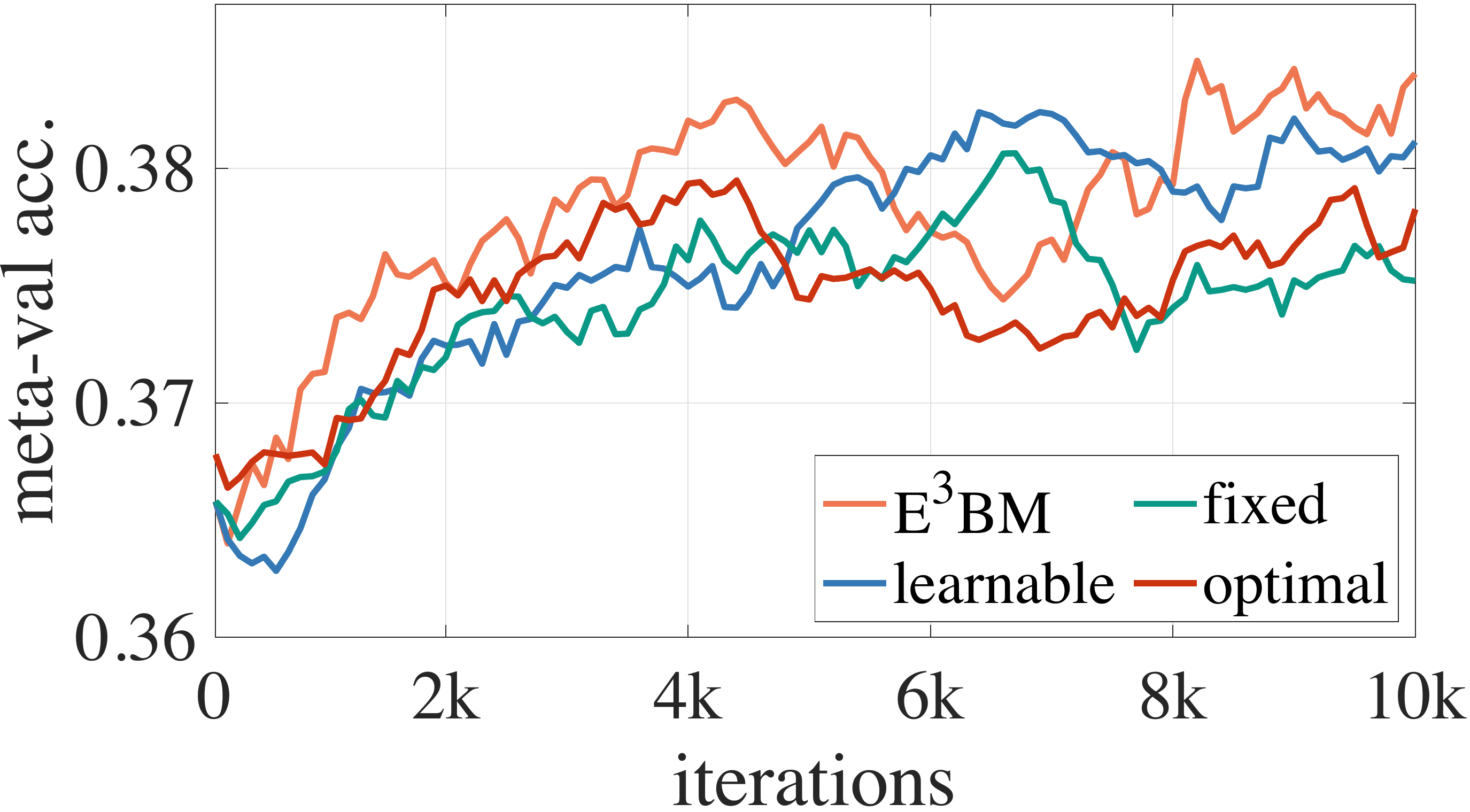}}

\subfigure[$v$, FC100, $5$-shot]{\includegraphics[height=1.15in]{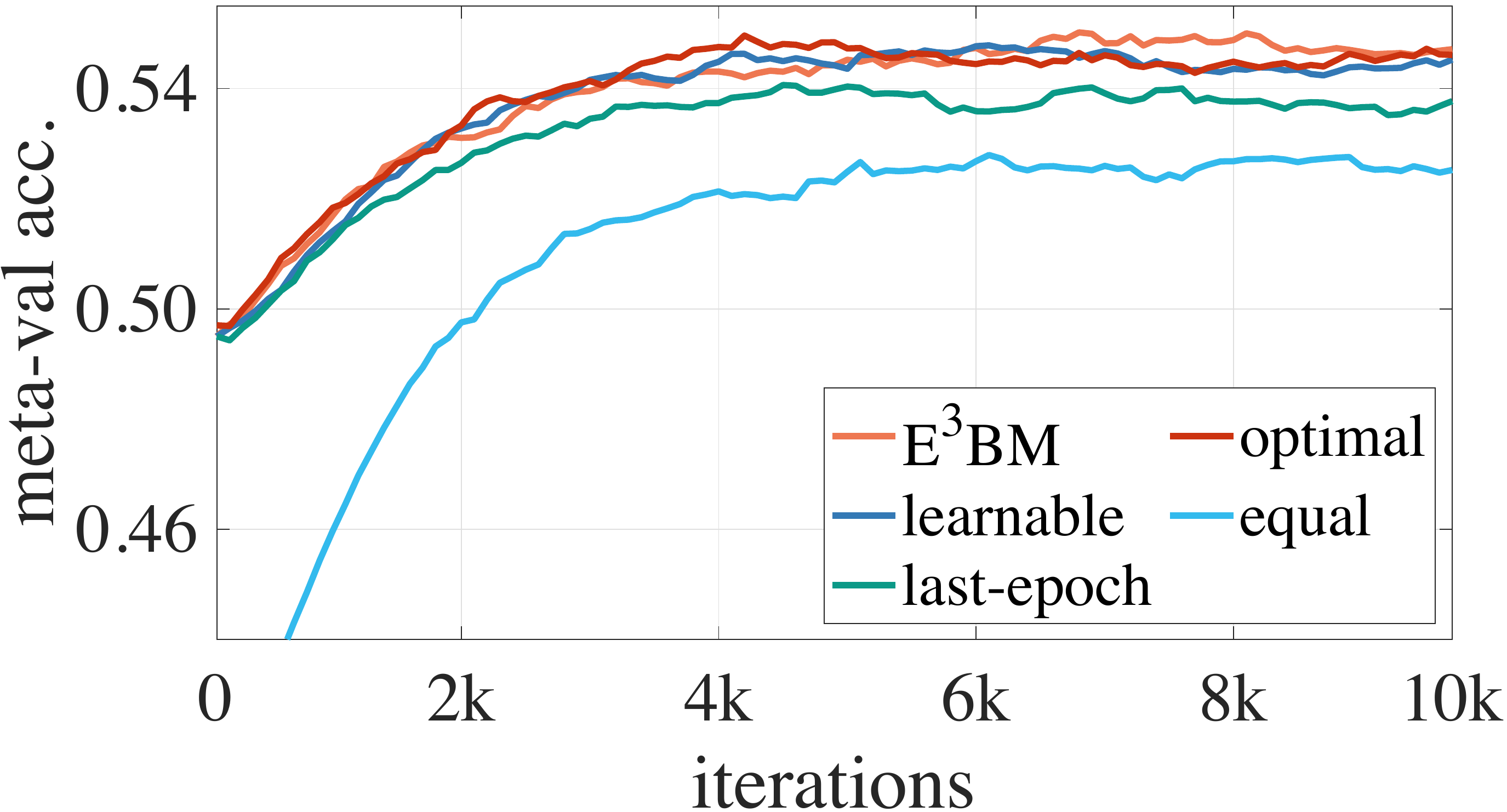}}
\ \ \ \ \ \ \ \ 
\subfigure[$\alpha$, FC100, $5$-shot]{\includegraphics[height=1.15in]{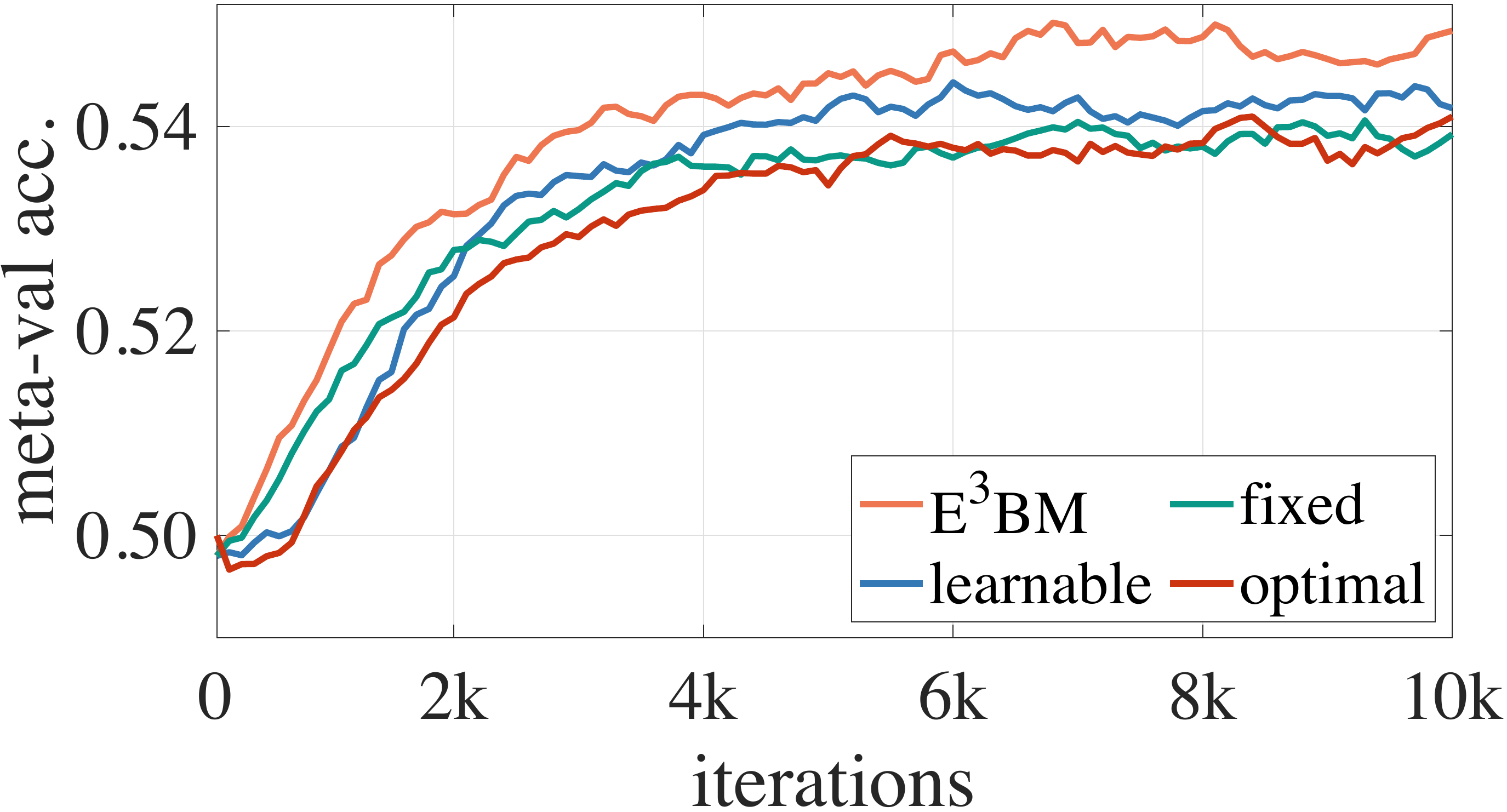}}

\end{center}

\caption{{\mycaptionsupp{Supplementary to Fig.\thinspace\ref{plot_mini_val_acc}(a)(b).}} The meta-validation accuracies of ablation models. Each figure demonstrates the results using the same model ``MTL+E$^3$BM'' as in Table 1. 
All curves are smoothed with a rate of $0.9$ for a better visualization.}

\label{plot_mini_5shot_val_acc}
\end{figure}

%% file: misc/5_all_value_alpha_v.tex
\begin{figure}
\begin{center}
\subfigure[$\alpha$, \emph{mini}ImageNet, $5$-shot]{\includegraphics[height=1.15in]{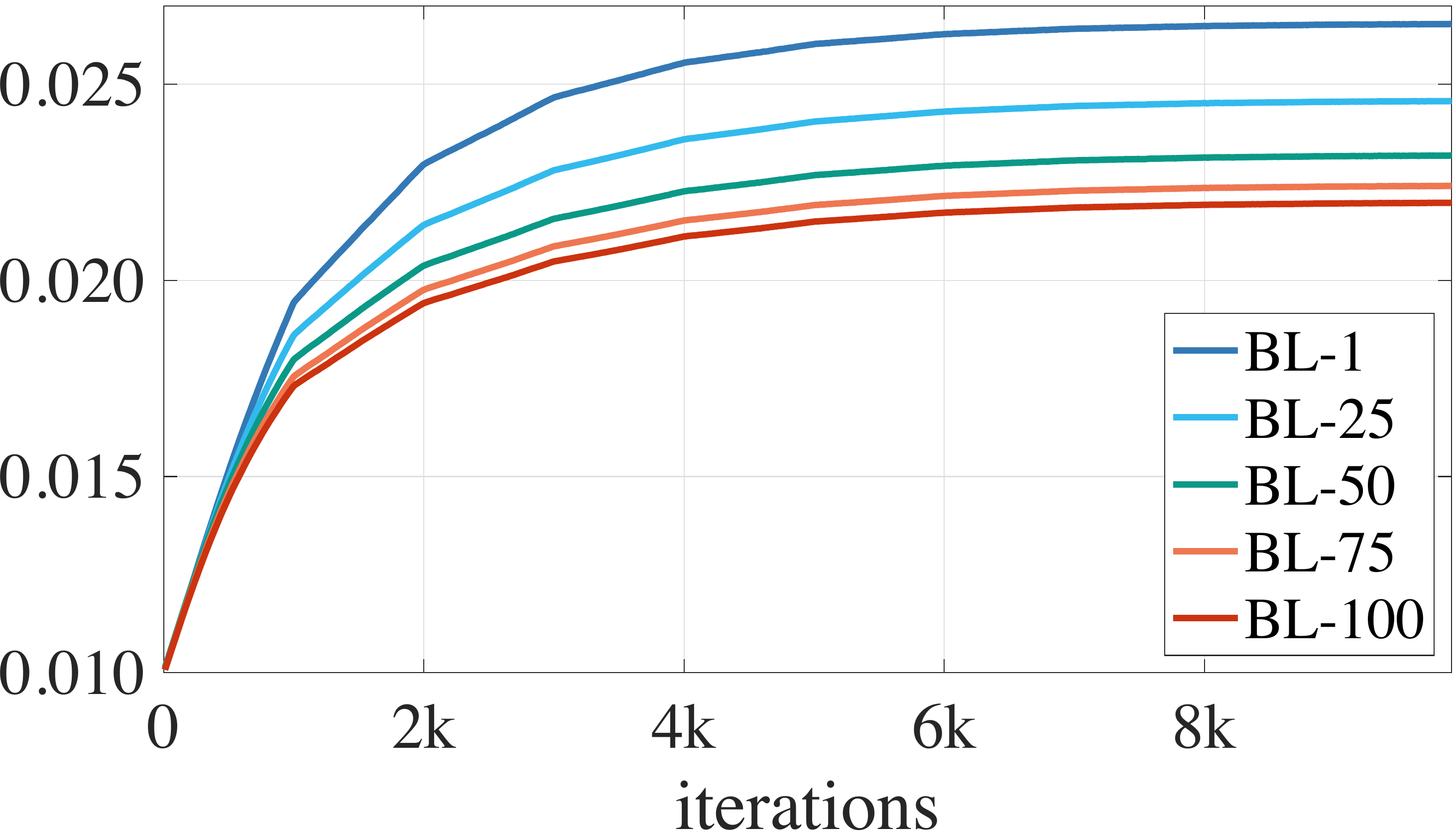}}
\ \ \ \ \ \ \ \ 
\subfigure[$v$, \emph{mini}ImageNet, $5$-shot]{\includegraphics[height=1.15in]{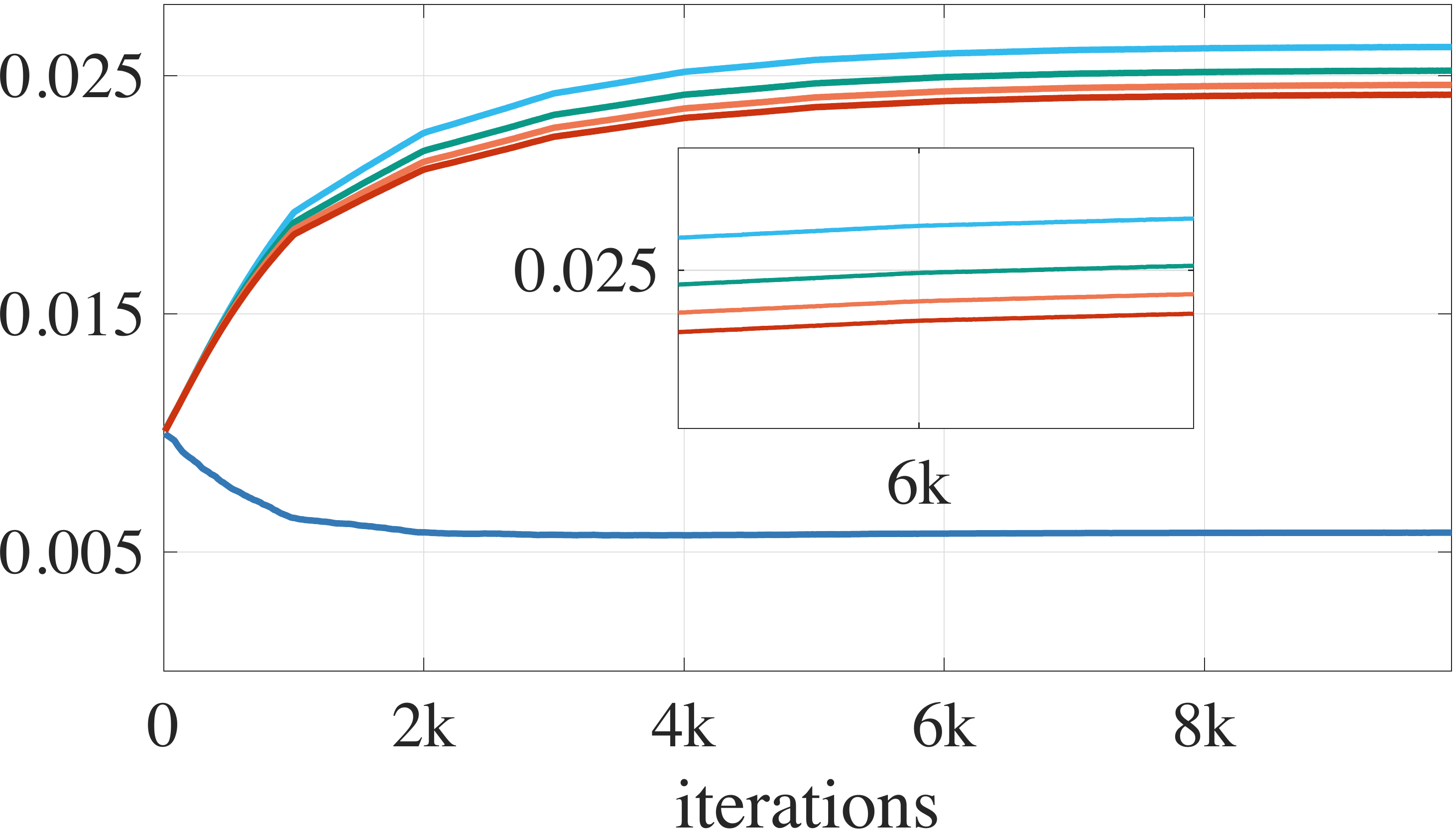}}

\subfigure[$\alpha$, \emph{tiered}ImageNet, $1$-shot]{\includegraphics[height=1.15in]{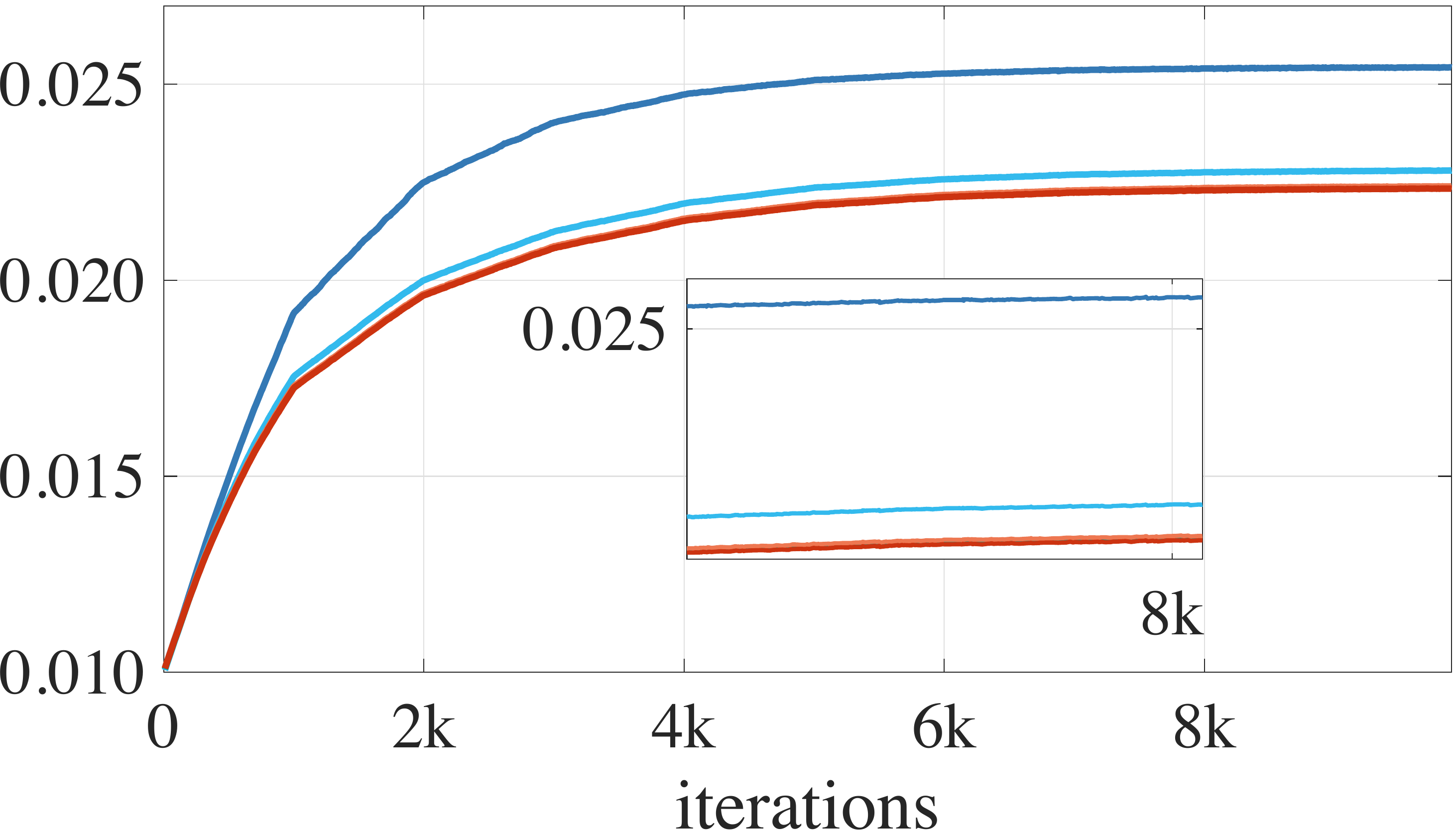}}
\ \ \ \ \ \ \ \ 
\subfigure[$v$, \emph{tiered}ImageNet, $1$-shot]{\includegraphics[height=1.15in]{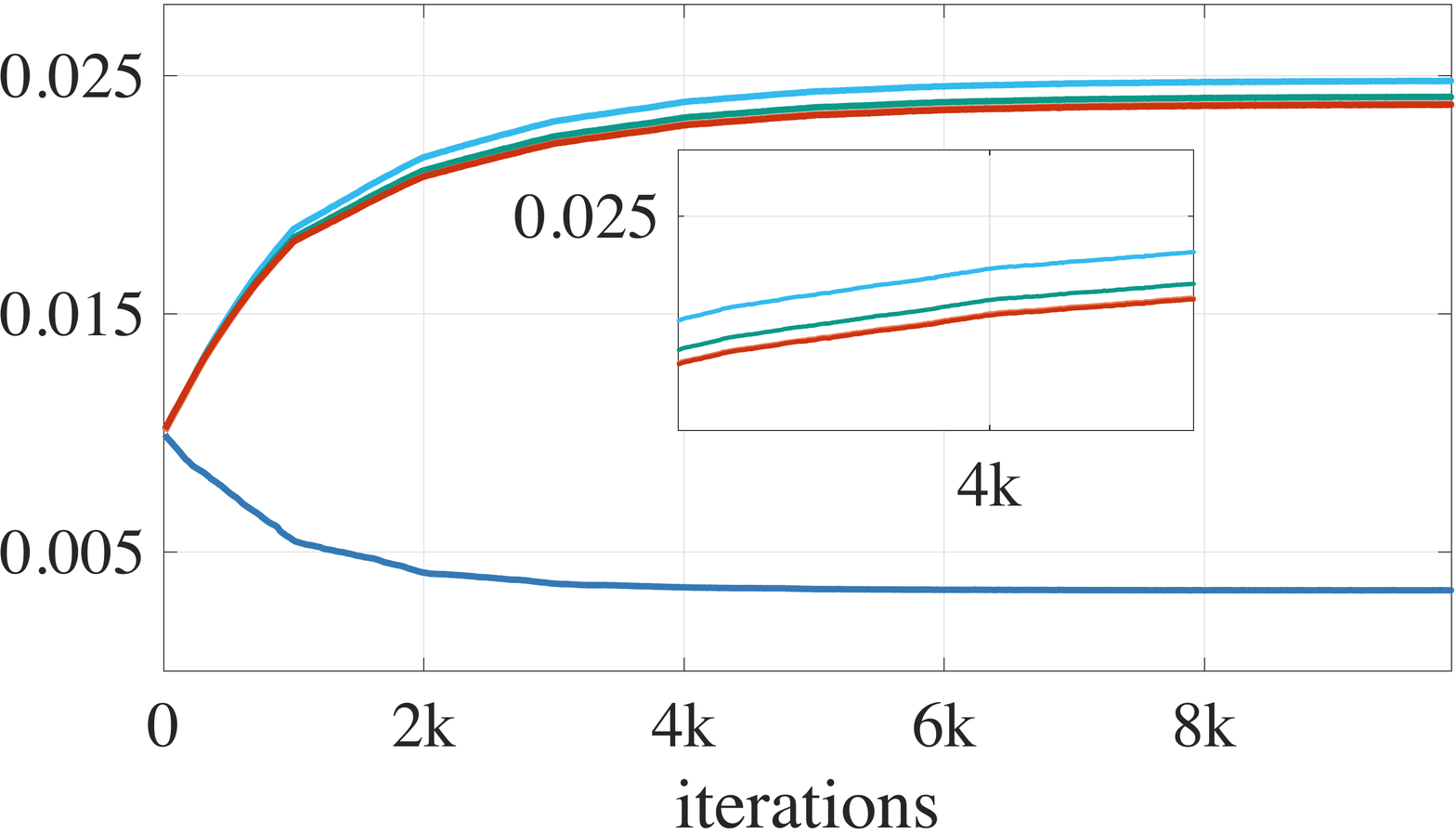}}

\subfigure[$\alpha$, \emph{tiered}ImageNet, $5$-shot]{\includegraphics[height=1.15in]{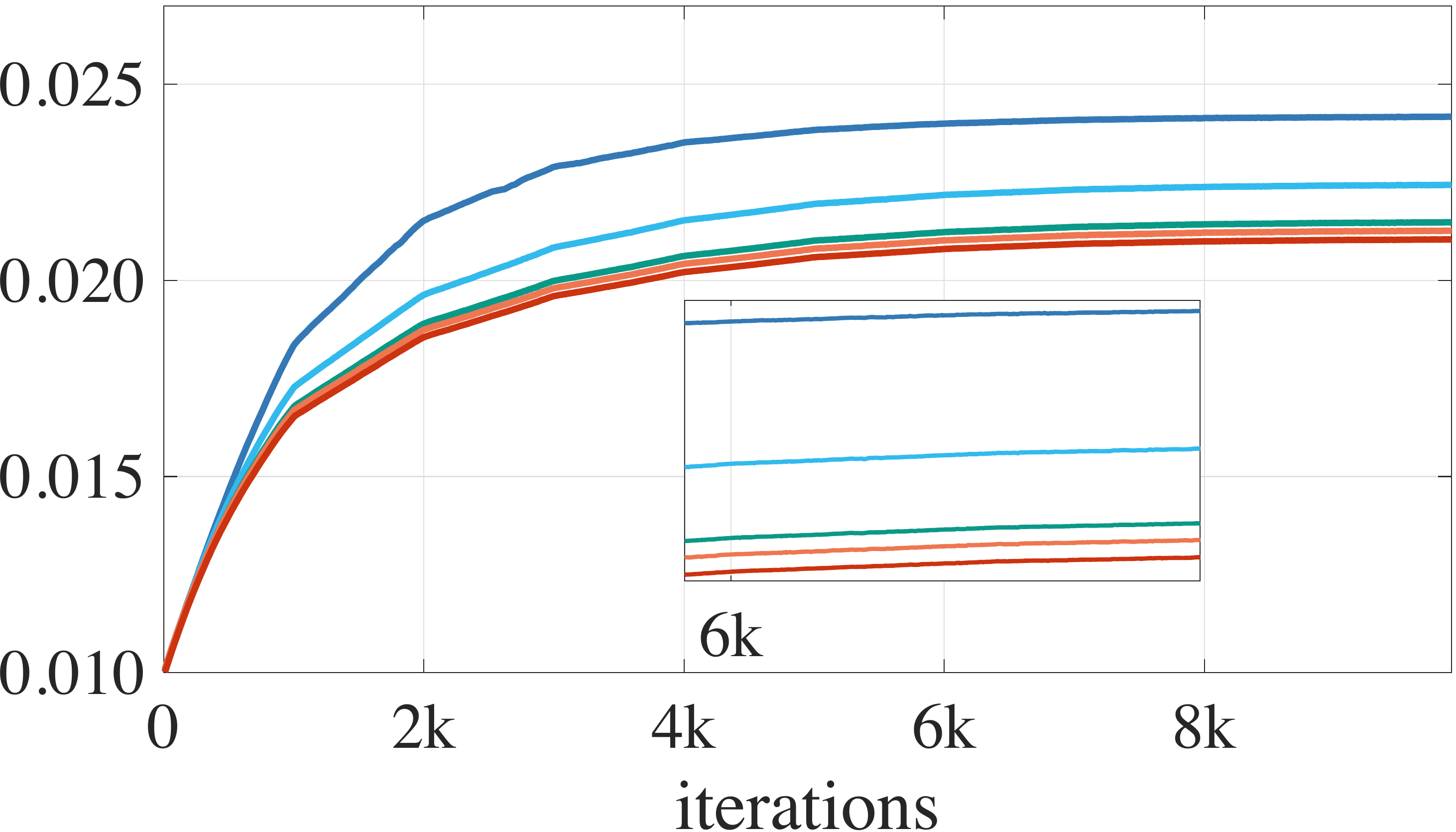}}
\ \ \ \ \ \ \ \ 
\subfigure[$v$, \emph{tiered}ImageNet, $5$-shot]{\includegraphics[height=1.15in]{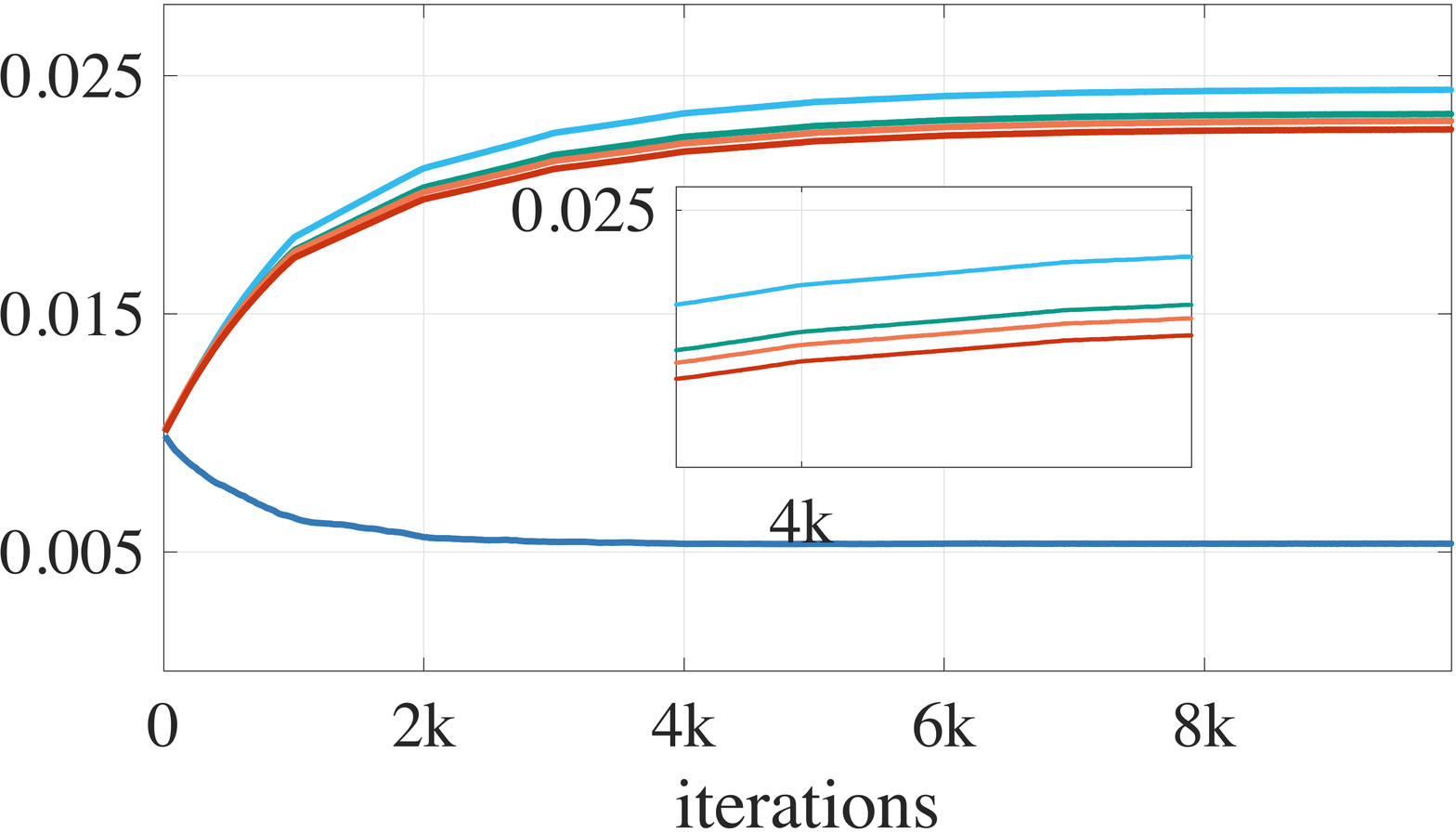}}

\subfigure[$\alpha$, FC100, $1$-shot]{\includegraphics[height=1.15in]{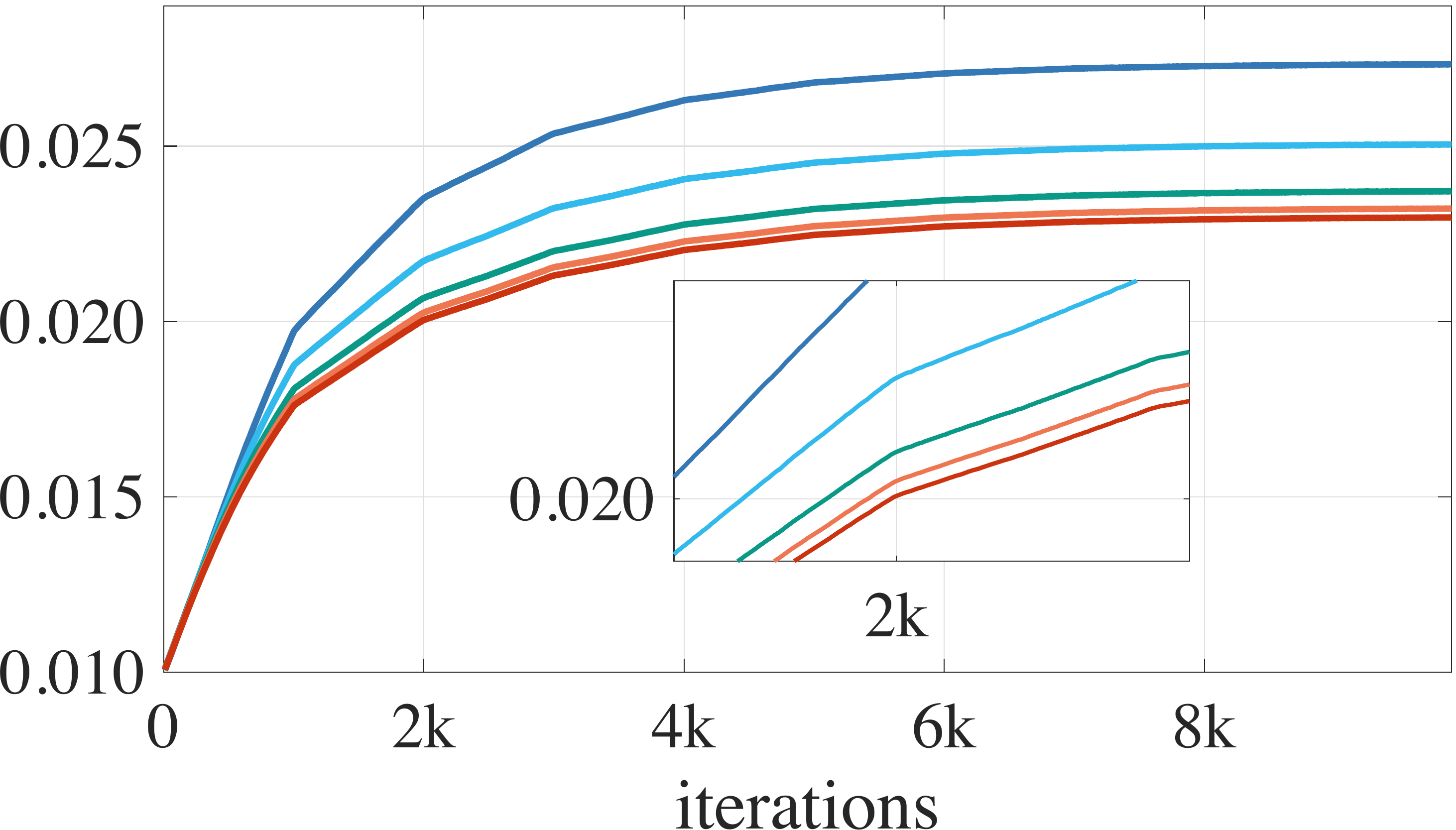}}
\ \ \ \ \ \ \ \ 
\subfigure[$v$, FC100, $1$-shot]{\includegraphics[height=1.15in]{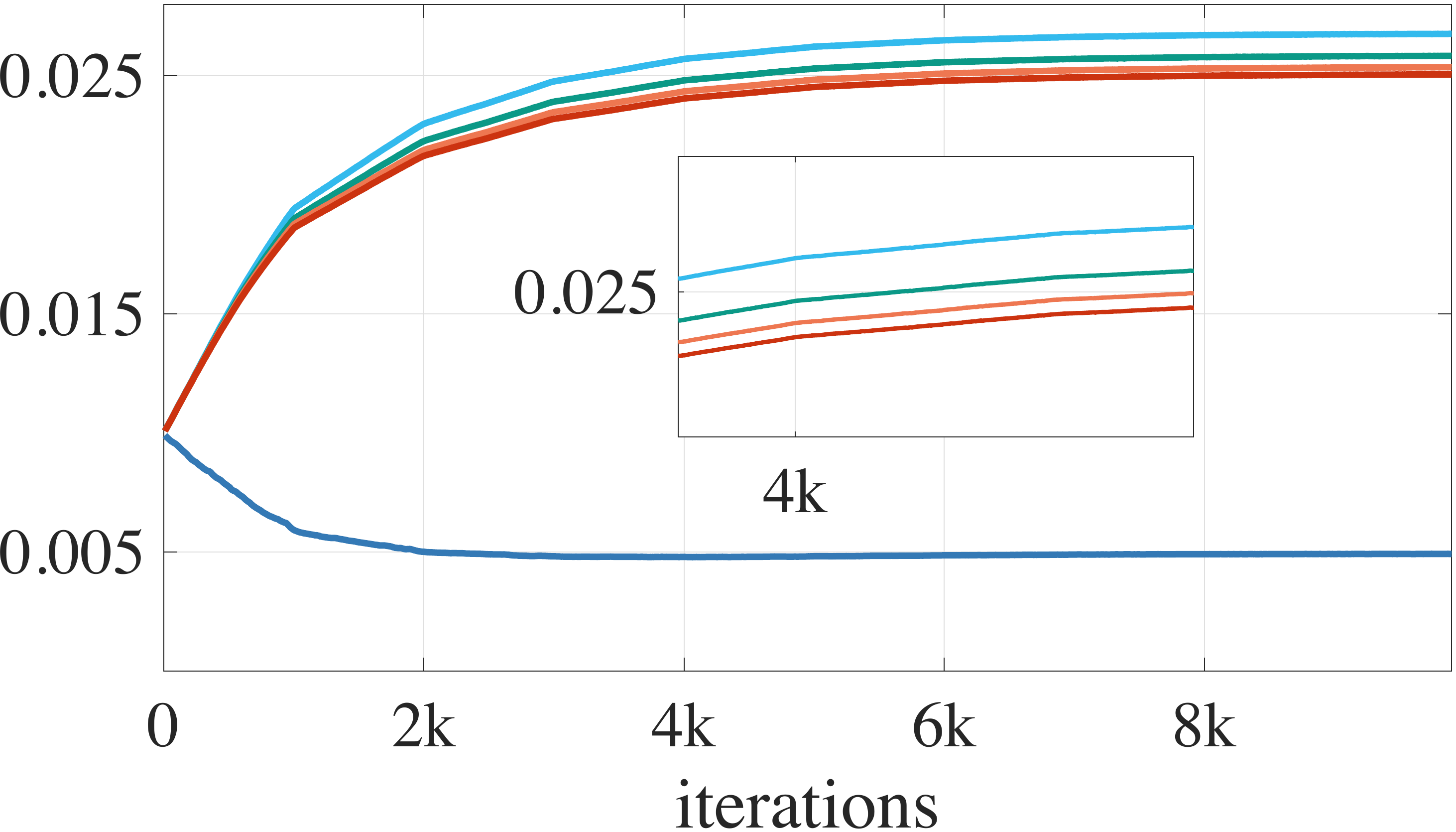}}

\subfigure[$\alpha$, FC100, $5$-shot]{\includegraphics[height=1.15in]{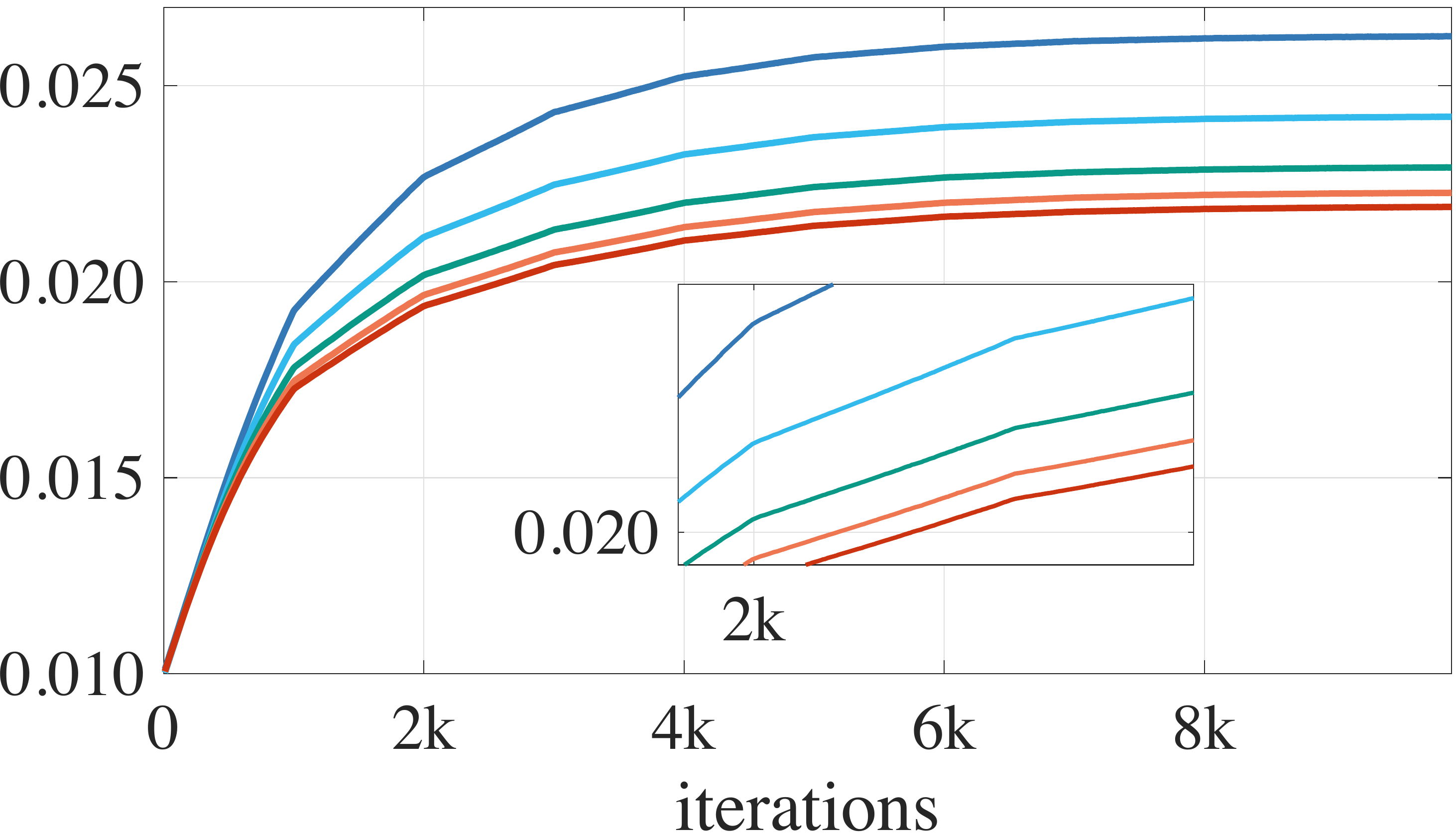}}
\ \ \ \ \ \ \ \ 
\subfigure[$v$, FC100, $5$-shot]{\includegraphics[height=1.15in]{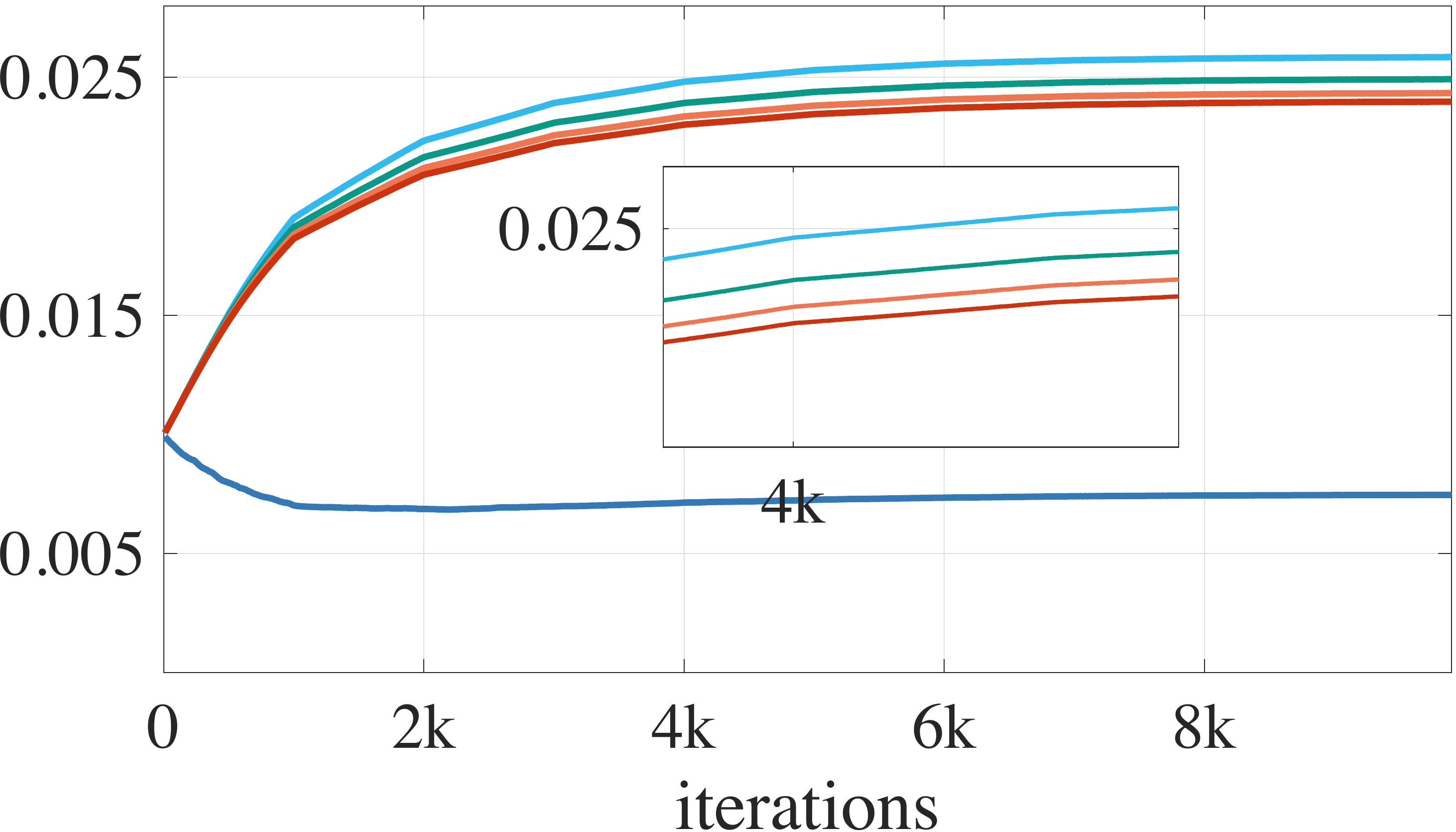}}

\end{center}

\caption{{\mycaptionsupp{Supplementary to Fig.\thinspace\ref{plot_mini_val_acc}(c)(d).}} The values of $\alpha$ and $v$ generated by $\Psi_{\alpha}$ and $\Psi_v$, respectively. Each figure demonstrates the results using the same model ``MTL+E$^3$BM'' as in Table 1. 
}

\label{plot_all_alpha_v}
\end{figure}